\documentclass[9pt,twocolumn,twoside]{osajnl}

\usepackage{mathtools}
\usepackage{bm}

\newcommand{\changed}[1]{#1}

\newcommand{\fref}[1]{Fig.~\ref{#1}}
\newcommand{\tref}[1]{Table~\ref{#1}}

\newcommand{\FloatBarrier}{}
\definecolor{Dark}{rgb}{0,0,0}

\journal{ol}
\setboolean{shortarticle}{false}
\dates{}

\newcommand{\scititle}{Resilient Odometry via Hierarchical Adaptation}
\title{\scititle}
\hypersetup{
  pdftitle={Resilient Odometry via Hierarchical Adaptation},
  pdfauthor={Shibo Zhao et al.}
}

\author{%
  \textcolor{color2}{\sffamily\bfseries
  Shibo Zhao$^{1,*}$, Sifan Zhou$^1$, Yuchen Zhang$^1$, Ji Zhang$^1$,
  Chen Wang$^2$, Wenshan Wang$^{1,\dagger}$, Sebastian Scherer$^{1,\dagger}$}\\[5pt]
  \small\sffamily\itshape
  $^1$Carnegie Mellon University, USA; $^2$University at Buffalo, USA\\
  $^*$Corresponding author: shiboz@andrew.cmu.edu;
  $^\dagger$Equally Advising\\
  Project Website: \href{https://superodometry.com}{superodometry.com}
}

\begin{abstract}
\textbf{Resilient and robust odometry is crucial for autonomous systems operating in complex and dynamic environments. 
Existing odometry systems often struggle with severe sensory degradations and extreme conditions such as smoke, sandstorms, snow, or low-light conditions, threatening both the safety and functionality of robots. To address these challenges, we present Super Odometry, a sensor fusion framework that dynamically adapts to varying levels of environmental degradation. 
Super Odometry employs a hierarchical structure to integrate four core modules from lower-level to higher-level adaptability including adaptive feature selection, adaptive state direction selection, adaptive engine selection, and a novel learning-based inertial odometry.
The inertial odometry, trained on over 100 hours of heterogeneous robotic platforms, captures comprehensive motion dynamics. 
Super Odometry elevates the inertial measurement unit (IMU) to equal importance with camera and LiDAR within the sensor fusion framework, providing a reliable fallback when exteroceptive sensors fail.
Super Odometry has been validated across 200 kilometers and 800 operational hours on a fleet of aerial, wheeled, and legged robots, under diverse sensor configurations, environmental degradation, and aggressive motion profiles. 
It marks an important step towards safe and long-term robotic autonomy in all-degraded environments.
}

\end{abstract}

\begin{document}
\maketitle
\thispagestyle{fancy}

% The class uses \doi for article metadata; the inline bibliography needs it for output.
\renewcommand{\doi}[1]{doi:\discretionary{}{}{}#1}

\section*{Summary}

A resilient odometry system that adapts seamlessly to diverse environments and robotic platforms.

\section*{Introduction}

% The first paragraph of any Science paper does NOT have a heading
% Nor is it indented
\noindent
Odometry is an important technique to estimate the position and orientation of robots over time, while also allowing for the 3D geometry reconstruction of surrounding environments. It plays a crucial role in robotics, enabling spatial understanding and serving as a foundation for both high-level tasks such as navigation and exploration, and low-level functions like path planning and control~\cite{thrun2002probabilistic}. As a result, odometry systems are widely used in robotic applications, including off-road driving~\cite{selfdriving}, search-and-rescue~\cite{Scherer:2022}, and planetary exploration~\cite{mahlknecht2022exploring}.

However, maintaining robust and resilient odometry performance in real-world applications remains challenging due to a range of environmental degradations, such as poor lighting, geometrical degradation, motion blur from aggressive maneuvers, and extreme weather. These conditions make it difficult for odometry systems to extract and track reliable features from sensors over time, often leading to incorrect data associations~\cite{cadena2016past} and eventual drift or failure.

To overcome these environmental degradations, researchers have explored multi-modal sensor fusion methods, implemented as either loosely-coupled~\cite{shen,wan2001unscented,zhang2014loam,zhang2015visual,complementaryodometry,camurri2020pronto} or tightly-coupled~\cite{graeter2018limo,zuo2019lic,wisth2021unified,r3live} frameworks. Although these approaches have substantially advanced the field, none can fully handle various forms of environmental degradation, particularly when visual and geometric signals degrade simultaneously~\cite{zhao2024subt}. This shortcoming continues to hinder the development of robust and
resilient autonomous navigation systems capable of long-term operation in diverse environments.
The key challenges can be summarized as follows:

\textit{Resilience Under Severe and Persistent Sensor Degradation:}
Existing odometry solutions face substantial challenges when multiple sensors experience degradation or remain off-nominal over extended periods. For example, persistent smoke in wildfire environments can affect both visual and LiDAR sensors, significantly hindering navigation~\cite{ebadi2023present}.

\textit{Balancing Robustness and Efficiency:}
Although many approaches improve robustness by integrating data from all available sensors to ensure redundancy and safety, they often introduce high computational costs due to complex fusion strategies~\cite{cadena2016past}. Achieving robustness without sacrificing efficiency remains an open challenge.

\textit{Generalization Across Varied Degradation Scenarios:}
Most methods need a significant amount of tuning for different types of environments since they are built on a ``rigid'' framework with a ``fixed'' parameter setting ~\cite{cadena2016past}. For example, a method optimized for outdoor conditions may fail in indoor environments such as subterranean environments. This lack of generalization limits their ability to generalize across diverse degradation conditions.

\begin{figure*}[!t]
    \centering
    \includegraphics[width=0.98\textwidth,keepaspectratio]{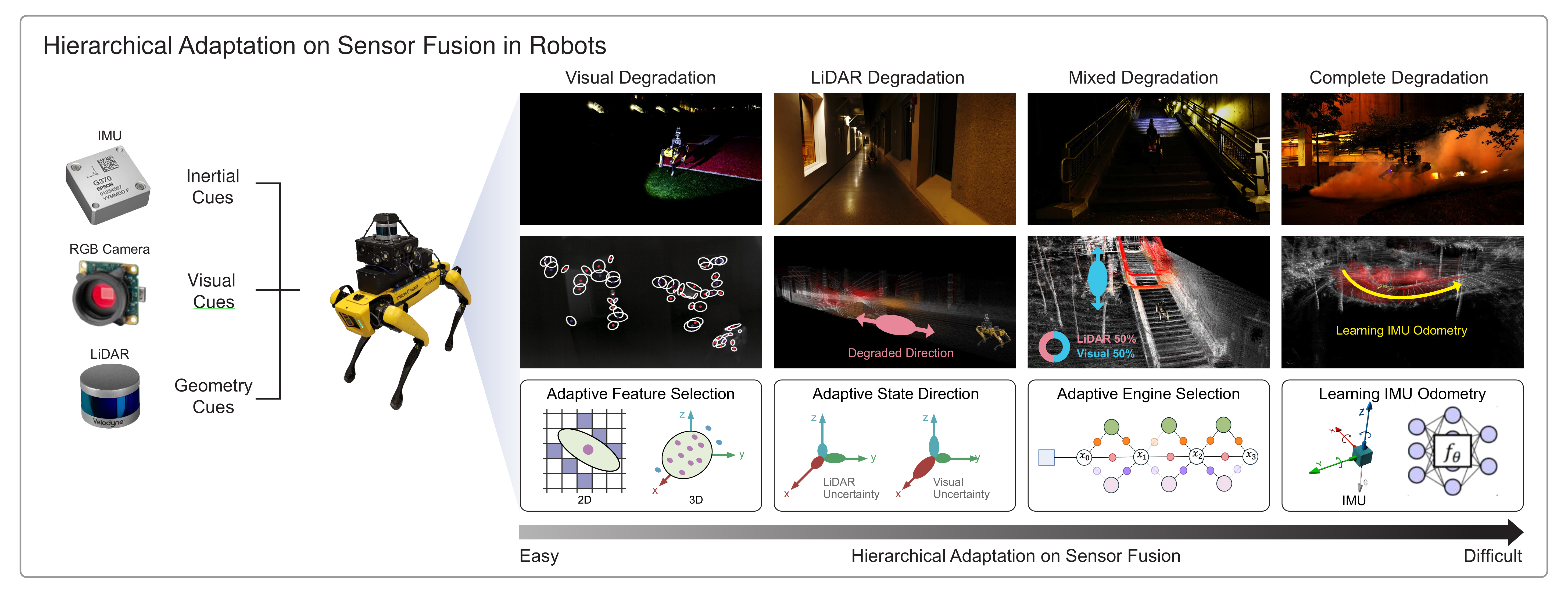}
    \caption{\textbf{Hierarchical Adaptation on Sensor Fusion in Robots.} \changed{We propose a hierarchical adaptation strategy that initiates with adaptive feature selection for visual degradation, moves to adaptive state direction for LiDAR degradation, applies adaptive factor (engine) selection for mixed degradation, and employs learning-based IMU odometry during complete degradation.}}
    \label{fig:concept}
\end{figure*}

To address the first challenge of resilience, we draw inspiration from the human sensing system. Humans navigate degraded environments such as dense fog or darkness by integrating external perception with internal cues. When vision becomes unreliable, they rely on vestibular and proprioceptive feedback to estimate motion, a process known as path integration~\cite{bostelmann2020path}. This allows them to maintain robust motion tracking even in the absence of external observation. Followed by this insight, we introduce a complementary internal sensing mechanism into robotic systems: a learned inertial module that infers motion from IMU data alone. Specifically, we develop a deep inertial network trained on diverse motion patterns to serve as an internal motion prior when external observations become unreliable.

To integrate this internal sensing into the odometry pipeline, we propose a reciprocal fusion strategy that combines traditional model-based odometry with the learned inertial module based on imperative learning \cite{wang2025imperative}. Instead of operating independently, these two components learn from each other: under nominal conditions, the traditional estimator refines the inertial network using accurate pose estimates, allowing it to adapt online. As environmental degradation worsens, the learned IMU network, having captured the system's motion dynamics, takes over estimation and ensures continued robustness. In this way, robustness becomes adaptive, evolving with the robot's operating conditions.

To address the second and third challenges, efficiency and generalization, we introduce a hierarchical adaptation framework that dynamically adjusts its behavior based on the severity of environmental degradation. The framework operates through a multi-level scheme: lower levels provide rapid and resource-efficient adaptations to address mild disturbances. If
these measures are insufficient, higher levels provide more complex and computationally intensive
interventions to support state estimation recovery. This layered design enables the system to maintain efficiency under nominal conditions and robustness under extreme scenarios, adapting both its computational load and sensing strategy to meet the demands of diverse environments.

In summary, we present Super Odometry, a sensor fusion framework designed for resilient, efficient, and generalizable state estimation under diverse and extreme sensor degradations. Our key contributions are as follows:

\textbf{Hierarchical Adaptation Mechanism:} 
We propose a hierarchical adaptive framework to address degradation along a spectrum from mild to severe and scale its response to maintain efficiency and robustness. It is dynamically reconfigurable, functioning as a multi-level scheme.  Specifically, the system initiates with adaptive feature selection to mitigate mild visual degradation, then transitions to adaptive state direction for moderate geometric degradation, proceeds to adaptive factor (engine) selection for mixed degradation,  and ultimately relies on a learning-based inertial odometry module when suffering from complete degradation, as illustrated in \fref{fig:concept}.

\textbf{Heterogeneous Learning-based Inertial Odometry:} 
Inspired by human path integration~\cite{bostelmann2020path}, we introduce a heterogeneous learning-based IMU odometry model to achieve internal motion estimation and predict 3D poses in real-time. Trained on hundreds of hours of diverse robotic data, it achieves strong generalization and maintains low-drift accuracy.

\textbf{Learning and Adaptation Over Time:}
We develop a self-supervised online adaptation algorithm for the data-driven inertial odometry using self-supervised imperative learning \cite{wang2025imperative,fu2024islam}. We formulate the system as a bilevel optimization: the IMU network learns motion patterns from free pose labels generated by a lower-level factor graph~\cite{dellaert2017factor}, while the upper-level inertial network provides motion priors back to the graph. This closed loop enables mutual correction between the learning-based and traditional estimators. For the first time, we elevate IMUs to an independent role in sensor fusion, positioning them as equally important as cameras and LiDAR systems.

\textbf{Extensive Evaluation:} Super Odometry was rigorously validated across over 200 kilometers and 800 operational hours over six years, on aerial, wheeled, legged, and handheld robots under diverse and extreme degradation, consistently outperforming state-of-the-art techniques.

We believe Super Odometry represents a notable step towards a resilient and robust sensor fusion solution capable of operating reliably anytime, anywhere, and in any degraded environment. We expect this advancement to enhance long-term robotic autonomy and ensure the safety of robots. For more details on motivation and methods, please refer to Supplementary Movie S1.

\FloatBarrier

% Research Articles and Reviews split the text into sections using headings
% Use a short (up 6 words) descriptive phrase, not generic 'Results' or 'Conclusions'
% Most other formats do not have headings, see the journal instructions to authors for details

\begin{figure*}[!t]
    \centering
    \includegraphics[width=0.9\textwidth,height=0.84\textheight,keepaspectratio]{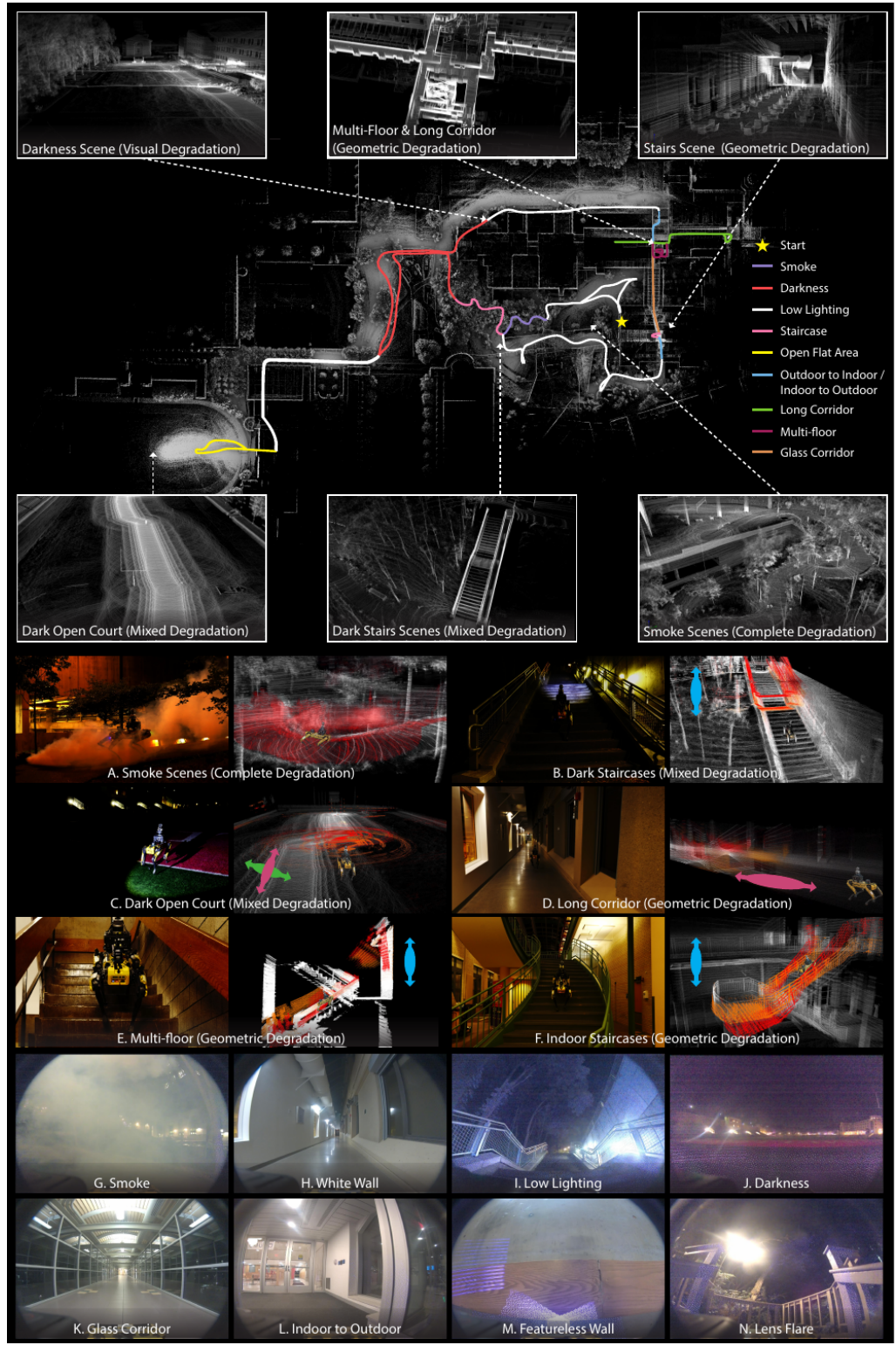}
    \caption{\textbf{Evaluation of 13 types of degradation in a single run.} The color-coded trajectory depicts our estimated odometry of a legged robot navigating through over 13 complex degradation scenarios. Despite these difficulties, the final endpoint drift was only \textbf{20 cm} over a total distance of 2,966 meters.}
    \label{fig:comprehensive_result}
\end{figure*}

\begin{figure*}[!t]
    \centering
    \includegraphics[width=\textwidth,keepaspectratio]{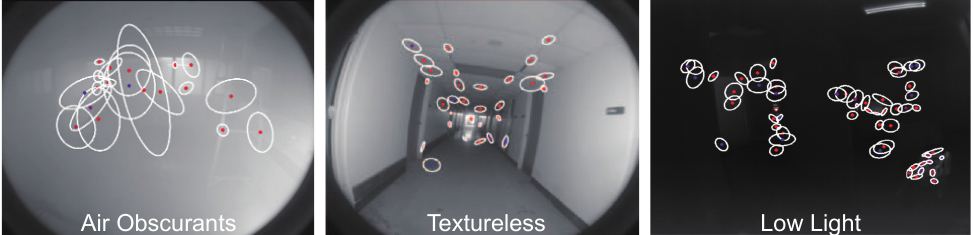}
    \caption{\textbf{Robust feature selection in low light, low texture, and obscured situations.} The ellipses represent the uncertainty of 2D feature landmarks; larger ellipse indicates greater uncertainty. The ellipse orientation reflects the ambiguity in feature tracking along a specific 2D direction.}
    \label{fig:visual_degradation}
\end{figure*}

\section*{Results}

Real-world environments are often complex, involving multiple types of degradation ranging from mild to extreme. Therefore, an odometry solution should be adaptable and adjustable.

We categorize degradation into four types:
Visual degradation refers to adverse visual conditions (e.g., low light, motion blur, and darkness) that impair feature extraction from cameras.
Geometric degradation arises in environments lacking meaningful 3D structure (e.g., open spaces and long corridors).
Mixed degradation occurs when visual and geometric cues degrade intermittently.
Complete degradation represents extreme scenarios where both visual and LiDAR sensors are off-nominal, such as in dense smoke.
 
For visual degradation, our approach relies on adaptive feature selection to reject outliers. In geometric and mixed degradation, it actively selects the state direction and switches between sensor engines to ensure the most reliable measurements are used. When facing extreme degradation such as complete degradation, our approach adopts a learning-based inertial odometry to overcome off-nominal sensor conditions such as smoke, dust, and snow.

In this section, we describe how our method performs across various degraded conditions. First, the method was evaluated through a comprehensive experiment, testing it under different degradation conditions in one run. Secondly, to illustrate the functionality of the method, we began with visual degradation as an example, highlighting the role of adaptive feature selection. Next, we used geometrically degraded environments to demonstrate adaptive state selection and mixed degraded environments for sensor engine switching. Finally, we explain how the learning-based IMU estimator helps overcome complete degradation in extreme smoke environments.

\subsection*{Comprehensive Degradation in One Run}

\paragraph{University Campus Environment}
To thoroughly evaluate the robustness of our odometry system, we conducted a comprehensive degradation experiment on a legged robot in a single run. This experiment included visual, geometrical, mixed, and complete degradation scenarios. The test route, located on Carnegie Mellon University's campus, covered a distance of 2966 meters and took approximately 46 minutes to complete. The route incorporates more than 13 complex degradation scenarios, involving various combinations of challenging conditions. These scenarios include geometrical degradations such as long, textureless corridors (Fig.~\ref{fig:comprehensive_result}D), narrow and steep multi-floor environments (Fig.~\ref{fig:comprehensive_result}E, F), and transitions between indoor and outdoor environments (Fig.~\ref{fig:comprehensive_result}L). Additionally, we introduced visual degradations (Fig.~\ref{fig:comprehensive_result}G-N) such as smoke, white walls, low lighting, darkness, glass corridors, featureless surfaces, and lens flare. Mixed degradations, like steep staircases in darkness (Fig.~\ref{fig:comprehensive_result}B) and expansive dark courts (Fig.~\ref{fig:comprehensive_result}C), were also tested. Finally, we examined complete degradation, where dense smoke enveloped the robot (Fig.~\ref{fig:comprehensive_result}A).

Our odometry system successfully completed real-time state estimation without any failures. The final endpoint drift was only 0.2 m over a total distance of 2966 meters, resulting in an exceptionally low drift rate of 0.006\%. The complete mapping results are shown in Fig.~\ref{fig:comprehensive_result}, highlighting high-fidelity views of the map under selected degraded conditions. The mapping results indicate the high accuracy and robustness of our pose estimates.  Notably, this performance was achieved without using any backend optimization techniques, such as loop closure. For more details on results, please refer to Supplementary Movie S2.

\FloatBarrier

\begin{figure*}[!t]
    \centering
    \includegraphics[width=0.62\textwidth,keepaspectratio]{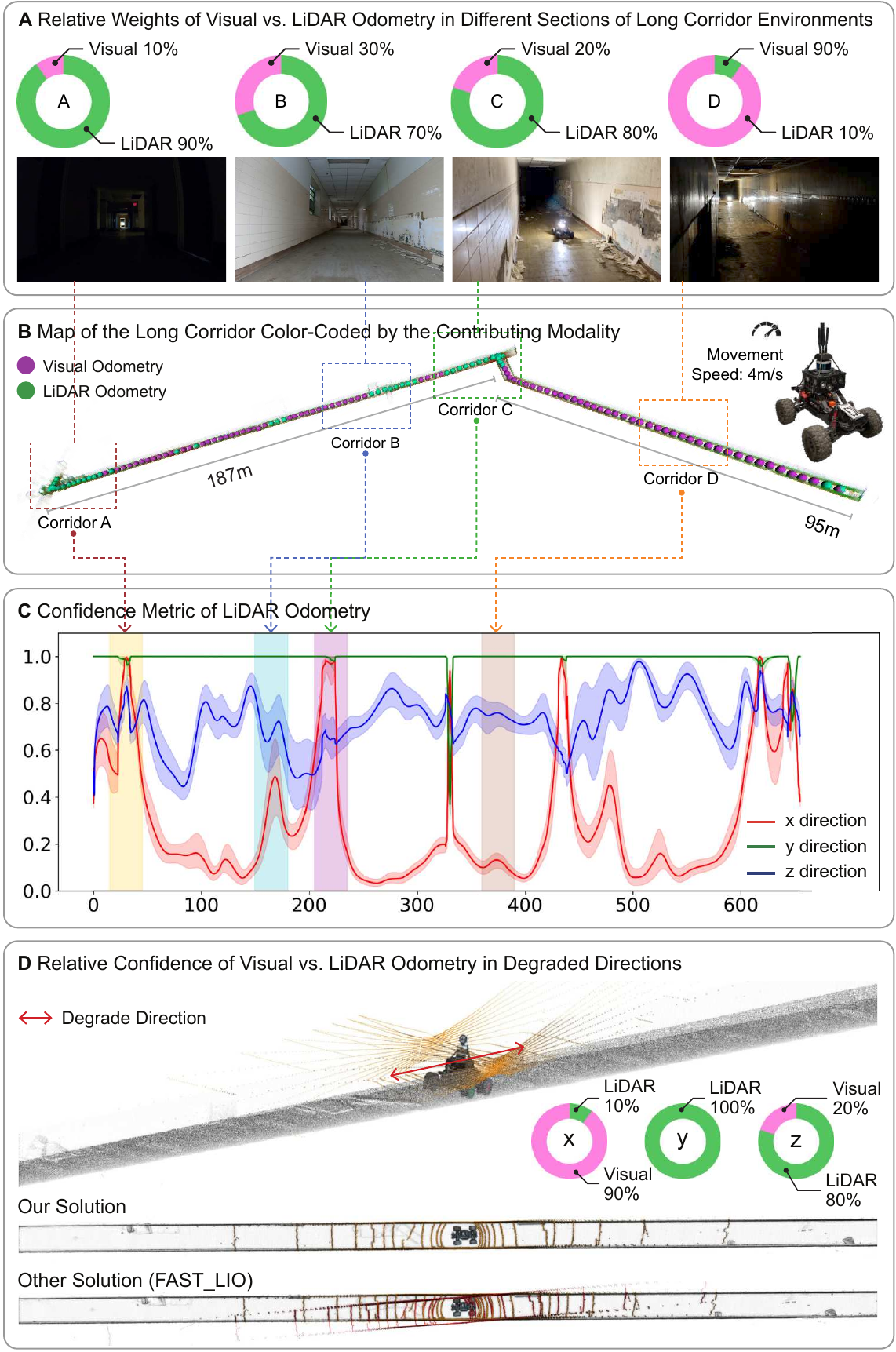}
    \caption{\textbf{Illustration of adaptive state direction selection in a long corridor \changed{from the abandoned Pittsburgh hospital}.} \textbf{(A)} Relative weights of visual and LiDAR odometry across different sections of long corridor environments. \textbf{(B)} Map of the long corridor, color-coded by the contributing modality; the purple circle indicates areas where visual odometry is actively integrated into sensor fusion, while the green circle denotes areas dominated by LiDAR odometry. \changed{Note: A modality is considered ``dominant'' if its weight in the sensor fusion process is greater than 50\%.} \textbf{(C)} Confidence metric of LiDAR odometry over a distance of 668 meters. \textbf{(D)} Relative confidence of visual and LiDAR odometry in the degeneracy direction, demonstrating more reliable scan registration achieved with our method.}
    \label{fig:long_corridor}
\end{figure*}

\subsection*{Adaptive Feature Selection for Visual Degradation}

In this section, we will primarily use visual degradation as an example to explain how adaptive feature selection works. We first conducted photometric calibration on RGB images by adjusting exposure time~\cite{bergmann2017online} which ensures that the images are consistent over time regardless of changes in lighting conditions. Then, we leveraged our previously proposed method ThermalPoint~\cite{zhao2020tp} to achieve robust feature detection on RGB images. Different from existing methods tracking all features 
passively~\cite{openvins,qin2018vins}, we took into account the quality of feature correspondences and estimated their covariance for each 2D feature. By doing this, our method can robustly select the most informative features in very challenging scenarios such as textureless corridors, low lighting conditions, and even air obscurants shown in \fref{fig:visual_degradation}. 

\FloatBarrier

\begin{figure*}[!t]
    \centering
    \includegraphics[width=0.95\textwidth,height=0.9\textheight,keepaspectratio]{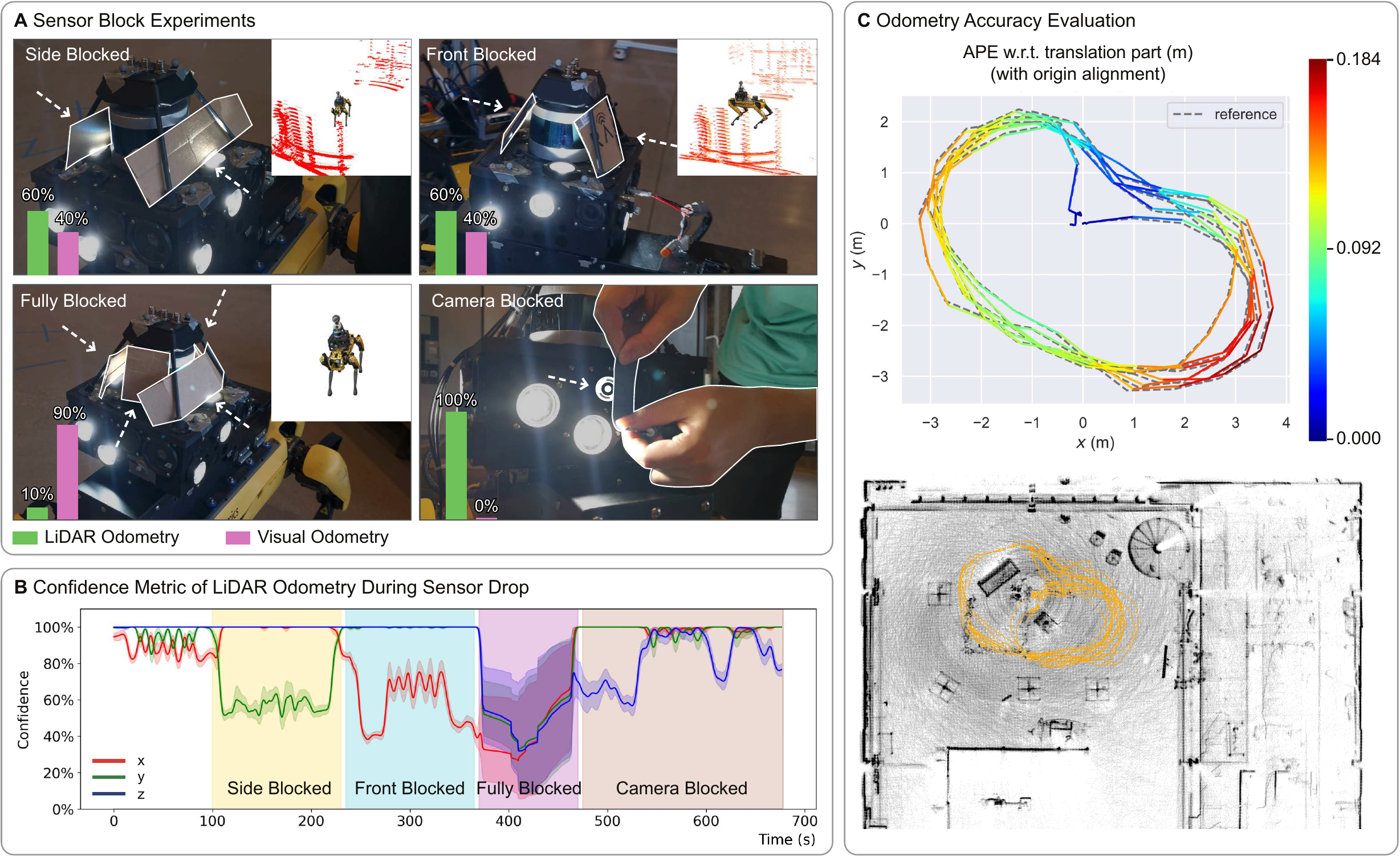}
    \caption{\textbf{Adaptive engine selection for mixed degradation in sensor-drop scenarios.} \textbf{(A)} We used cardboard to sequentially block the LiDAR's side, front, and back views, and eventually completely obstructed both the LiDAR and visual sensor to evaluate the dropout tolerance for 600 seconds. \textbf{(B)} The corresponding confidence metric of LiDAR Odometry System. \textbf{(C)} Odometry accuracy evaluation: the trajectory color encodes APE error (top), and the bottom shows mapping results with yellow indicating the estimated trajectory.}
    \label{fig:drop_sensor}
\end{figure*}

\subsection*{Adaptive State Direction for Geometric Degradation}

% \paragraph{Long Corridor} 
In this section, we will use long corridors to illustrate the adaptive state direction mechanism. Long corridors, with repetitive and symmetrical features such as uniform walls lacking distinct geometry,
present substantial challenges for LiDAR systems~\cite{xu2021fast}. The similarity of these features across different positions makes Iterative Closest Point (ICP) registration~\cite{segal2009generalized}
poorly constrained in the forward direction.  

The experiment involved navigating a completely dark structureless corridor with highly reflective walls (Fig.~\ref{fig:long_corridor}A). Despite these difficulties, we successfully completed the testing route of 668 meters. The vehicle operated aggressively, maintaining a speed of 4 m/s for up to 90\% of the route and reaching a maximum speed of 5 m/s (Fig.~\ref{fig:long_corridor}B).

To achieve this, we developed an adaptive state direction method, which actively predicted the degeneracy direction of the optimization and incorporated other pose priors (visual) to mitigate the ill-optimization problem before failure. Fig.~\ref{fig:long_corridor}C shows the confidence metric of LiDAR Odometry reliability on X, Y, and Z directions in the long corridor environments.   It is observed that the confidence in the X direction, corresponding to the robot-centered forward direction, is relatively low compared to those in the Y and Z directions during the run. 

Fig.~\ref{fig:long_corridor}D illustrates the relative confidence levels of visual and LiDAR odometry in the degeneracy direction. Our method assigned lower confidence to LiDAR odometry in the forward direction, whereas visual odometry maintained higher confidence. The weights for LiDAR odometry in the Y and Z directions remain unchanged, as there was no degradation in those axes. Additionally, our method outperformed state-of-the-art methods like FAST-LVIO\cite{fastlivo} in achieving reliable scan registration in poorly constrained long corridors environments.

By analyzing the relative confidence of LiDAR and visual odometry, we color-code the trajectory on the LiDAR map based on each sensor's contribution, as shown in Fig.~\ref{fig:long_corridor}B. In the corridor's middle section, visual odometry was actively integrated into sensor fusion and addresses degeneracies, whereas at the end section, the system increasingly relies on LiDAR odometry. This process employed soft fusion between LiDAR and visual odometry across each axis, providing additional constraints at optimal moments. For more results, please refer to Supplementary Movie S3.

\FloatBarrier

\subsection*{Adaptive Engine Selection for Mixed Degradation}

Real-world environments present various challenges, where systems can experience visual or geometric degradation at different stages, classified as mixed degradation.  A typical example was sensor-drop scenarios, where either visual or LiDAR data may be compromised and suffer from interruptions for extended periods. To solve this, we believe an odometry solution should be fail-safe and failure-aware. It should detect imminent failures like signal loss or degradation and provide recovery mechanisms.

We initially blocked the LiDAR's side view using the cardbocard shown in Fig.~\ref{fig:drop_sensor}A. Our algorithm, adopting an adaptive state-direction mechanism, promptly predicted the degradation in the Y direction and reduced the LiDAR confidence weight to 60\% in that direction (Fig.~\ref{fig:drop_sensor}B). Simultaneously, the visual odometry confidence weight was increased to 40\% to compensate for the LiDAR's reduced performance. This process not only predicted the degraded direction but also adapted the necessary weights and constraints between the visual and LiDAR modality to prevent optimization failure. 

Similarly, when the front and back view of the LiDAR was blocked, the LiDAR's confidence weight in the X direction dropped to 60\%, whereas the weight of the visual odometry in that direction increased to 40\%. Upon fully blocking the LiDAR with cardboards, its weight in all directions dropped to 10\%, whereas the visual odometry's confidence weight rose to 90\%. Finally, when we blocked the camera, the LiDAR's confidence weight across all directions returned to over 90\%. Importantly, if a modality's contribution to the joint state fell below 10\% and remained low for more than 2--4 seconds, the system will disable that modality. This threshold and temporal window were empirically chosen and tunable to ensure fail-safe and failure-aware operation. Fig.~\ref{fig:drop_sensor}C illustrates the odometry and mapping accuracy to evaluate the sensor dropout tolerance of our method.  The largest Absolute Trajectory Error (ATE) is only 0.184m over a 600-second run.

\FloatBarrier
\begin{figure*}[!t]
    \centering
    \includegraphics[width=0.82\textwidth,height=0.9\textheight,keepaspectratio]{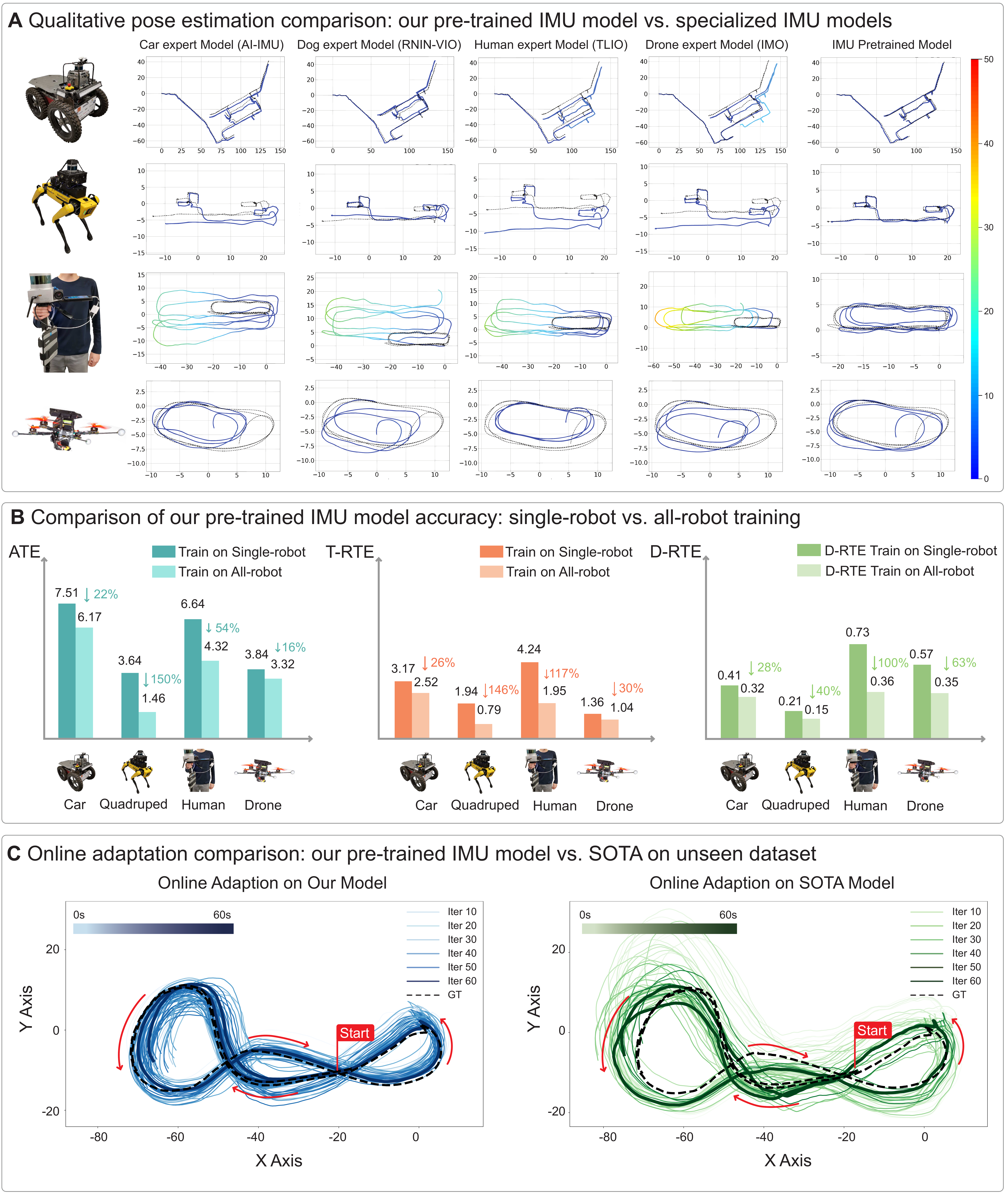}
    \resizebox{0.74\linewidth}{!}{%
        \tiny
        \begin{tabular}{lccccccc}
        \hline
        \changed{Timeline} & \changed{0 s} & \changed{10 s} & \changed{20 s} & \changed{30 s} & \changed{40 s} & \changed{50 s} & \changed{60 s} \\
        \hline
        \changed{SOTA Model ATE (m)} & \changed{34.14} & \changed{29.24} & \changed{18.20} & \changed{15.56} & \changed{13.28} & \changed{9.48} & \changed{4.94} \\
        \changed{Our Model ATE (m)} & \changed{\textbf{32.87}} & \changed{\textbf{27.06}} & \changed{\textbf{13.49}} & \changed{\textbf{8.27}} & \changed{\textbf{5.34}} & \changed{\textbf{4.03}} & \changed{\textbf{1.73}} \\
        \hline
        \changed{$\uparrow$ Improve ($\%$)} & \changed{3.72} & \changed{8.06} & \changed{25.88} & \changed{46.86} & \changed{59.79} & \changed{57.49} & \changed{64.98} \\
        \hline
        \end{tabular}
    }
    \caption{\textbf{Evaluation on learning-based inertial odometry.} \textbf{(A)} Qualitative pose comparison shows our method (last column) achieves the highest accuracy against Ground Truth (dashed gray). \textbf{(B)} Accuracy improves significantly when the IMU pre-trained model is trained on multi-robot data versus single-robot data. \textbf{(C)} Online adaptation on unseen datasets demonstrates our model outperforms SOTA, \changed{as supported by a quantitative comparison of ATE over time. Each iteration shown corresponds to 1s of adaptation time (e.g., Iteration 10 represents 10 seconds).}}
    \label{fig:FIMU_ablation}
\end{figure*}

\begin{figure*}[!t]
    \centering
    \includegraphics[width=0.64\textwidth,height=0.95\textheight,keepaspectratio]{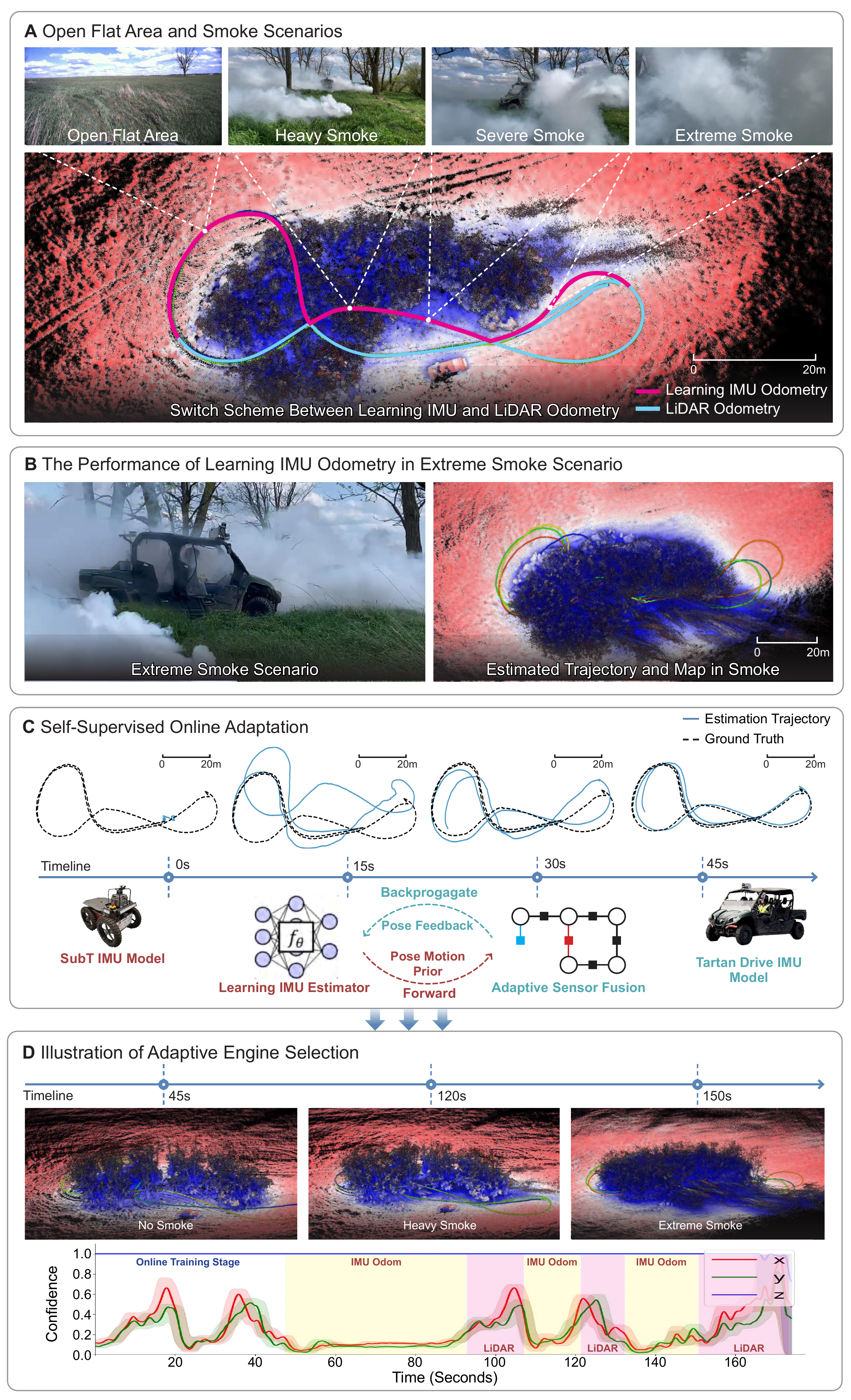}
    \caption{\textbf{Robust Performance in a Smoke Scenario.} \textbf{(A)} Adaptive engine selection between learning-based IMU and LiDAR odometry in challenging environments, with purple and blue trajectories indicating their respective use. \textbf{(B)} Trajectory and map generated using learning-based IMU odometry in heavy smoke. \textbf{(C)} Self-supervised adaptation fine-tunes the IMU model from ``SubT car'' to ``Tartan Drive.'' \textbf{(D)} Confidence-driven adaptive fusion combines IMU and LiDAR odometry outputs using soft fusion, avoiding hard switches.}
    \label{fig:offroad_smoke}
\end{figure*}

\begin{table*}[!t]
\caption{\textbf{Quantitative comparison of pose estimation between the IMU pre-trained model and specialized IMU models.} Our IMU pre-trained model outperformed different specialized IMU models, achieving an average of 35.5\% on ATE and an average of 41\% on T-RTE, respectively, compared to the second-best model across various robotic platforms, including wheeled, legged, handheld, and aerial systems.}
\vspace{2pt}
\centering
\sffamily
\setlength{\tabcolsep}{2pt}
\renewcommand{\arraystretch}{1.12}
\resizebox{\textwidth}{!}{
\begin{tabular}{c|cc|cc|cc|cc|cc|cc}
    \hline
    % \multirow{2}{*}{}
    \textbf{Robot Platform}
    & \multicolumn{2}{c|}{\textit{AI-IMU}~\cite{AI-IMU}} 
    & \multicolumn{2}{c|}{\textit{RNIN-VIO}~\cite{chen2021rnin}} 
    & \multicolumn{2}{c|}{\textit{TLIO}~\cite{liu2020tlio}} 
    & \multicolumn{2}{c|}{\textit{IMO}~\cite{IMO}} 
    & \multicolumn{2}{c|}{\textit{IMU-Pretrain (Ours)}}  
    & \multicolumn{2}{c}{\textit{Improvement}} \\ 
    \cline{2-13}
    & ATE $\downarrow$ & T-RTE $\downarrow$ 
    & ATE $\downarrow$ & T-RTE $\downarrow$ 
    & ATE $\downarrow$ & T-RTE $\downarrow$ 
    & ATE $\downarrow$ & T-RTE $\downarrow$ 
    & ATE $\downarrow$ & T-RTE $\downarrow$ 
    & ATE & T-RTE \\ 
    \hline
    Wheeled (Car)~\cite{zhao2024subt} 
    & 7.68 & 3.33 & 7.82 & 5.06 & 8.12 & 3.73 & 8.12 & 3.73 & \textbf{6.17} & \textbf{2.52} & \changed{\textbf{$\uparrow$ 19.63\%}}  & \changed{\textbf{$\uparrow$ 24.36\%}}   \\ 
    Legged (Quadruped)~\cite{zhao2024subt} 
    & 3.23 & 1.60 & 3.10 & 1.58 & 3.61 & 1.73 & 3.35 & 1.64 & \textbf{1.46} & \textbf{0.79} & \changed{\textbf{$\uparrow$ 54.83\%}}   & \changed{\textbf{$\uparrow$ 50.63\%}}   \\ 
    Handheld (Human)~\cite{sun2021idol} 
    & 8.26 & 4.89 & 7.62 & 5.61 & 6.96 & 4.82 & 10.19 & 5.67 & \textbf{4.32} & \textbf{1.95} & \changed{\textbf{$\uparrow$ 47.69\%}}   & \changed{\textbf{$\uparrow$ 60.05\%}}   \\ 
    Aerial (Drone)~\cite{antoniniIJRRblackbird} 
    & 4.14 & 1.45 & 4.32 & 1.51 & 3.93 & 1.40 & 3.72 & 1.34 & \textbf{3.32} & \textbf{1.04} & \changed{\textbf{$\uparrow$ 19.81\%}}   & \changed{\textbf{$\uparrow$ 28.97\%}}   \\ 
    \hline
\end{tabular}
}
\label{tab:learning_imu_odom}
\end{table*}

\subsection*{Learning Based Inertial Odometry for Complete Degradation}

In degraded environments like dense smoke, dust storms, and heavy snow,  almost all the odometry algorithms~\cite{shan2020lio,xu2021fast,fastlivo} face severe challenges, as they encounter both visual and geometric degradation over an extended period. Visual sensors struggle to capture high-quality images needed to extract reliable features. LiDAR systems, which depend on laser pulses for distance measurements, experience reduced range and accuracy due to air obscurants. 

To address these limitations, we propose a learning-based IMU odometry since IMU is unaffected by perceptual degradation. Our solution is straightforward: when other sensors, such as visual or LiDAR, are functioning well, the IMU odometry receives feedback from them and fine-tunes its path integration capabilities~\cite{bostelmann2020path}. However, when these sensors are compromised, such as in extreme environments, the IMU odometry takes over state estimation.

\paragraph{IMU Pre-trained Model on Large Dataset}
Developing robust IMU odometry is challenging, especially in ensuring generalization across platforms. Most methods lack adaptability beyond specific settings~\cite{IMO}. We addressed this by proposing a general IMU model trained on large, diverse datasets~\cite{zhao2024subt, sivaprakasam2024tartandrive}, covering drones, quadrupeds, cars, and human. This broad training enabled it to outperform specific models~\cite{chen2018ionet,liu2020learning,chen2021rnin,yan2018ridi} and adapt to various platforms. Our approach aimed at strong generalization, rapid adaptation, and seamless integration into sensor fusion pipelines. To validate these objectives, we investigated three core research questions:
\textbf{Q1:} Can our model generalize across platforms and outperform specialized models?
\textbf{Q2:} Can it adapt online to previously unseen environments?
\textbf{Q3:} How does it enhance sensor fusion robustness in extreme environments?

\paragraph{Generalization of IMU-Pretrained Model}

To address \textbf{Q1}, we focused on two critical factors for training a more generalized IMU model: a heterogeneous shared backbone and large, high-quality, diverse data.

To demonstrate the importance of a heterogeneous shared backbone, we evaluated different IMU models on data from four platform types: aerial, wheeled, legged, and handheld. For a fair comparison, all models, including AI-IMU~\cite{AI-IMU}, RNIN-VIO~\cite{chen2021rnin}, TLIO~\cite{liu2020tlio}, and IMO~\cite{IMO}, were trained on the same large datasets and tested on identical unseen data.

As shown in \fref{fig:FIMU_ablation}(A), our IMU pre-trained model, built with a heterogeneous shared backbone, demonstrated strong few-shot capabilities. When tested across the four robotic platforms, it consistently outperformed specialized models tailored for specific systems. Without requiring fine-tuning, our model provided the most reliable and accurate odometry estimates across a variety of motion patterns (see \fref{fig:FIMU_ablation}(A) last column). On average, it improved ATE and Time-Relative Trajectory Error (T-RTE) by 35.5\% and 41.0\%, respectively, compared to the second-best model (see Table \ref{tab:learning_imu_odom}). These results underscore the effectiveness of the heterogeneous shared backbone in achieving robust generalization across diverse robotic motion patterns.

We also investigated the effect of data diversity on model performance. Specifically, we compared the performance of our IMU pre-trained model when trained on data from a single platform versus data from various robot types. As illustrated in Fig.~\ref{fig:FIMU_ablation}(B), models trained on diverse datasets showed substantial performance improvements. For instance, in the ``human handheld'' sequence, a model trained on data from aerial, wheeled, legged, and quadruped platforms outperformed one trained exclusively on human handheld data. The ATE decreased by up to 54\%, whereas T-RTE and Distance-Relative Trajectory Error (D-RTE) decreased by 117\% and 100\%, respectively.
These findings confirm the critical role of large, high-quality, and diverse datasets in improving model generalization. Training on a wide variety of robot data enables the model to perform well even when there are substantial differences between the training and testing domains.  For more details on results, please refer to Supplementary Movie S4.

\paragraph{Fast Online Adaptation on \changed{Test Data from TartanDrive}}

\changed{Although our model was trained on a large dataset (as described in \textbf{Q1}), it still faces generalization issues when deployed on a new robot platform with different motion patterns. Moreover, collecting sufficient data for fine-tuning during deployment is often impractical. To address \textbf{Q2}, we use online fine-tuning during deployment---we learn as we operate.}

We assessed our model's ability to generalize to an unseen platform. The model was trained exclusively on the SubT dataset, which features a small ground vehicle with a maximum speed of 5 m/s. To evaluate its adaptability, we tested it on the TartanDrive dataset, which represents a full-size all-terrain vehicle with a maximum speed of 15 m/s. During the online fine-tuning process, the model was provided with only 60 seconds of IMU data, enabling it to adapt in real-time from the ``SubT UGV Model'' to the ``TartanDrive Car Model.''

We compared our model against a state-of-the-art baseline model~\cite{chen2021rnin}, trained on the same SubT dataset as ours, following an identical training process. In Fig.~\ref{fig:FIMU_ablation}(C), over a 60-second fine-tuning period, our model (blue lines) demonstrated significantly better odometry predictions, closely aligning with the ground truth (dashed line). In contrast, the state-of-the-art model (green lines), although showing some improvement, exhibited substantial trajectory errors at the 60-second mark.

\changed{This online adaptation scenario shows the potential of our pre-trained IMU model, which can adapt to new domains more rapidly and converges with fewer iterations than the state-of-the-art methods, as detailed in the table from Fig.~\ref{fig:FIMU_ablation}(C)}.

\FloatBarrier

\paragraph{Enhance Odometry Robustness in Smoke Environments} To answer \textbf{Q3}, \changed{we designed experiments where the robot is required to operate in a smoke-filled environment} to demonstrate how learning-based IMU odometry can improve the robustness of odometry systems. It included two processes: self-supervised online adaptation (explained in \textbf{Q2}) and learning-based IMU model deployment in the sensor fusion pipeline.

\textit{Self-supervised Online Adaptation: }
As discussed in \textbf{Q2}, Fig.~\ref{fig:offroad_smoke}(C) illustrates the self-supervised online adaptation process used to fine-tune our pre-trained IMU model from the SubT vehicle to the TartanDrive vehicle. Initially, the TartanDrive vehicle operated in a smoke-free environment, where our factor graph pipeline continuously provided supervisory signals to refine the learning-based IMU odometry. This online training process required only 45 seconds of data collection to successfully \changed{evolve} the pre-trained ``SubT'' IMU model into the ``TartanDrive'' model.

\textit{Learning-based Inertial Odometry Deployment: } 
After fine-tuning the IMU model, we integrated the learning-based IMU odometry into our adaptive sensor fusion pipeline and evaluated its performance online using data \changed{from human-driven vehicles}. We evaluated its performance in two challenging environments: an extremely smoke-filled area and a clear, open flat area, as depicted in Fig.~\ref{fig:offroad_smoke}A. In these environments, LiDAR odometry became highly unreliable. When the vehicle entered the smoke or open flat areas, our adaptive pipeline automatically prioritized the learning-based IMU odometry solution (purple trajectory). This is because the learning-based IMU odometry remained unaffected by these environmental conditions and continued to provide reliable state estimation. Conversely, when the vehicle exited these areas and conditions improved, the pipeline seamlessly transitioned back to the LiDAR odometry solution (blue trajectory). This switching behavior was governed by the confidence level of the LiDAR odometry, as illustrated in Fig.~\ref{fig:offroad_smoke}D.

Importantly, this transition was not a \changed{``hard switch''} but a soft fusion approach that blended the outputs of both LiDAR and IMU odometry across each axis. This strategy mirrored the adaptive engine selection mechanism used in previous mixed degradation scenarios.

To validate our method, we conducted figure-eight driving tests, repeatedly transitioning between smoky and clear areas. The adaptive switching occurred multiple times, demonstrating the pipeline's responsiveness to environmental changes. The estimated trajectory and the corresponding map are shown in Fig.~\ref{fig:offroad_smoke}B. Although the reconstructed map exhibits severe noise due to the smoke, our sensor fusion pipeline consistently delivered robust state estimation.

This experiment demonstrated that incorporating learning-based IMU odometry significantly enhances the robustness of odometry systems in severely degraded environments. For more details on results, please refer to Supplementary Movie S5.

\FloatBarrier

\begin{table*}[!t]
\caption{\textbf{ATE performance on SubT-MRS \cite{Zhao2024CVPR}.} * denotes incorporation of loop closure. - denotes incomplete trial.}
\centering
\sffamily
\setlength{\tabcolsep}{3pt}
\renewcommand{\arraystretch}{1.12}
\resizebox{0.72\textwidth}{!}{
\begin{tabular}{@{}c@{\hspace{3pt}}|ccccc|ccc|c@{}}
% \toprule
\hline
% \multirow{2}{*}{\textbf{Method}} & \multicolumn{5}{c|}{\color{Dark} \textbf{Geometric Degradation}} &\multicolumn{3}{c|}{\color{Dark} \textbf{Mixed Degradation}} & \multirow{2}{*}{\textbf{Average}} \\
% \cmidrule(lr{0.5em}){2-6}  \cmidrule(lr{0.5em}){7-9}

\textbf{Method} & \multicolumn{5}{c|}{\color{Dark} \textbf{Geometric Degradation}} &\multicolumn{3}{c|}{\color{Dark} \textbf{Mixed Degradation}} & \textbf{Average} \\
% \cmidrule(lr{0.5em}){2-6}  \cmidrule(lr{0.5em}){7-9}
\hline
 
 & \color{Dark} F01  & \color{Dark} F02 & \color{Dark} F03 & \color{Dark} U01  & \color{Dark} U02  & \color{Dark} {Cave03}   & \color{Dark}Corridor01 & \color{Dark}Floor01 & \\ 
% \cmidrule(lr){1-1} \cmidrule(lr){2-2} \cmidrule(lr){3-3} \cmidrule(lr){4-4} \cmidrule(lr){5-5} \cmidrule(lr){6-6} \cmidrule(lr){7-7} \cmidrule(lr){8-8} \cmidrule(lr){9-9} \cmidrule(lr){10-10}
\hline
Liu\text{*}~\cite{Xu2022,liu2023efficient}& 0.307  & 0.095 & 0.629 & 0.122 & 0.235 & 0.260      & 1.454 & 0.401  & 0.588\\
Weitong\text{*}~\cite{wu2025dali,dellaert2017factor}  & 0.26 & 0.096 & 0.617 & 0.120 & 0.222 & 0.402 & 1.254 & 0.577 & 0.663 \\
Kim\text{*}~\cite{xu2021fast,lim2022quatro} & 0.331          & 0.092         & 0.787      & 0.123        & 0.270 & 0.279  & 2.100   & 0.650 & 3.825  \\
Yibin ~\cite{wu2024icra}& 1.060          &0.220         & 0.750           & 0.470         & 0.620 & 9.140          & 2.990 & 5.500 & 4.312  \\
Zhong\text{*}~\cite{chen2022direct,kim2018scan}   & 1.205 & 0.695 & - & 1.175 & 1.72 & 2.08 & -  & - & 1.209  \\
Zheng~\cite{chen2022direct}  & - & - & - & - & - & 3.786 & 55.205  & 19.769 & 26.254  \\
Our  & \textbf{0.238} & \textbf{0.074} & \textbf{0.396} & \textbf{0.026} & \textbf{0.104} & \textbf{0.204} & \textbf{0.817} & \textbf{0.315} & \textbf{0.271}  \\
% \midrule
\hline
\end{tabular}
}

\label{tab:AccLidar}
\vspace{-4mm}
\end{table*}

\begin{table*}[!t]
\caption{\textbf{Robustness performance on SubT-MRS \cite{Zhao2024CVPR}.} * denotes loop closure usage. - denotes incomplete trial.}
\vspace{2pt}
\centering
\sffamily
\setlength{\tabcolsep}{2pt}
\renewcommand{\arraystretch}{1.1}
\resizebox{0.78\textwidth}{!}{
\begin{tabular}{@{}l|c|ccccc|ccc|c@{}}
% \toprule
\hline
\multicolumn{2}{c|}{} &\multicolumn{5}{c|}{\color{Dark} \textbf{Geometric Degradation (SubT-MRS)}} &\multicolumn{3}{c|}{\color{Dark} \textbf{Mix Degradation}}&\\
% \cmidrule(lr{0.5em}){3-7}  \cmidrule(lr{0.5em}){8-10} 
\hline
 
\multicolumn{2}{l|}{\textbf{Method}} & \color{Dark} Final01  & \color{Dark} Final02 & \color{Dark} Final03 & \color{Dark} Urban01  & \color{Dark} Urban02  & \color{Dark} {Cave03}   & \color{Dark}Corridor01 & \color{Dark}Floor01 & \textbf{Average}\\ 
% \cmidrule(lr){1-2}  \cmidrule(lr){3-3} \cmidrule(lr){4-4} \cmidrule(lr){5-5} \cmidrule(lr){6-6} \cmidrule(lr){7-7} \cmidrule(lr){8-8} \cmidrule(lr){9-9} \cmidrule(lr){10-10} \cmidrule(lr){11-11}
\hline
% Weitong\text{*}\cite{Bai2022,dellaert2017factor}  & \multirow{7}{*}{Rp} & 0.922 & 0.929 & 0.906 & 0.933 & 0.919 & 0.830 & 0.889 & 0.909 & 0.905  \\
Weitong\text{*}\cite{wu2025dali,dellaert2017factor}  & {Rp} & 0.922 & 0.929 & 0.906 & 0.933 & 0.919 & 0.830 & 0.889 & 0.909 & 0.905  \\
Liu\text{*}\cite{Xu2022,liu2023efficient}  && 0.920 & 0.927 & 0.904 & 0.931 & 0.917 & 0.832 & 0.885 & 0.905 & 0.903\\
Kim\text{*}\cite{xu2021fast,lim2022quatro} & & 0.884 & 0.929 & 0.906 & 0.933 & 0.915 & 0.830 & 0.890 & 0.252 & 0.817  \\
Yibin\cite{wu2024icra} & & 0.849 & 0.897 & 0.827 & 0.875 & 0.795 & 0.751 & 0.502 & 0.738 & 0.779    \\
Zhong\text{*}~\cite{chen2022direct, kim2018scan} & & 0.278 & 0.910 & 0.827 & 0.905 & 0.877 & - & - & - & 0.759 \\
Zheng\cite{chen2022direct}  & & - & - & - & - & - & 0.618 & 0.579 & 0.389 & 0.529 \\
Our&  &  \textbf{0.929} & \textbf{0.933} & \textbf{0.916} & \textbf{0.937} & \textbf{0.925} & \textbf{0.973} & \textbf{0.899} & \textbf{0.891} & \textbf{0.925}\\
% \midrule
\hline
% Weitong \text{*}\cite{Bai2022,dellaert2017factor} & \multirow{7}{*}{Rr}& 0.931 & 0.931 & 0.907 & 0.938 & 0.924 & 0.848 & 0.890 & 0.914 & 0.911\\
Weitong \text{*}\cite{wu2025dali,dellaert2017factor} & {Rr}& 0.931 & 0.931 & 0.907 & 0.938 & 0.924 & 0.848 & 0.890 & 0.914 & 0.911\\
Liu\text{*}\cite{Xu2022,liu2023efficient} && 0.930 & 0.930 & 0.908 & 0.937 & 0.922 & 0.848 & 0.885 & 0.908 & 0.908\\
Kim\text{*}\cite{xu2021fast,lim2022quatro} & & 0.923 & 0.931 & 0.908 & 0.938 & 0.920 & 0.848 & 0.890 & 0.260 & 0.827  \\
Yibin\cite{wu2024icra} & & 0.924 & 0.928 & 0.898 & 0.933 & 0.914 & 0.846 & 0.874 & 0.886 & 0.900\\
Zhong\text{*}~\cite{chen2022direct,kim2018scan}& & 0.296 & 0.921 & 0.851 & 0.922 & 0.905 & - & - & - & 0.779\\
Zheng\cite{chen2022direct}  & & - & - & - & - & - & 0.629 & 0.882 & 0.489 & 0.667 \\
Our& &  \textbf{0.938} & \textbf{0.942} & \textbf{0.925} & \textbf{0.943} & \textbf{0.941} & \textbf{0.964} & \textbf{0.944} & \textbf{0.924} & \textbf{0.940}\\
% \midrule
\hline
\end{tabular}
}
\label{tab:RobustLidar}
\end{table*}

\subsection*{Quantitative Comparison with State-of-the-Art Methods}

\paragraph{Accuracy Evaluation} To further validate the precision of pose estimation, we conducted ATE\cite{sturm2012benchmark} analysis using our odometry system on the SubT-MRS dataset \cite{Zhao2024CVPR}. This dataset encompasses challenging environments featuring sensor degradation, aggressive locomotion, and extreme weather conditions. The eight sequences in the dataset are categorized into two groups for testing: Geometric degradation and Mixed degradation. ATE results of competing systems were sourced from an open odometry challenge \cite{Zhao2024CVPR}. Participating teams were allowed to employ loop closure techniques and offline processing to achieve high accuracy. \changed{Despite this, our method achieved an average ATE of 0.271 in real-time.} This performance surpassed the second-best method by 54\%, showing our high accuracy and robustness in Table~\ref{tab:AccLidar}.

\paragraph{Robustness Evaluation} 
An ideal evaluation metric should assess both accuracy and completeness \cite{Zhao2024CVPR}. Although the ATE effectively measures trajectory accuracy, it does not adequately capture trajectory completeness. To address this limitation, we employed robustness metrics \cite{zhao2021super}, which account for both aspects. Table~\ref{tab:RobustLidar} presents a detailed comparison of these metrics across various scenarios and algorithms, highlighting that our method demonstrates superior robustness compared to competing approaches.

It is important to note that this comparison excludes completely degraded or smoke-related experiments (as illustrated in Fig.~\ref{fig:offroad_smoke}). The exclusion was due to the fact that all other algorithms fail under such conditions. For details on the robustness metric, please refer to the Method section.

\FloatBarrier

\section*{Discussion}
In this section, we provide insights \changed{on developing a robust odometry for degraded environments}.

\paragraph{Hierarchical Adaptation is a Key Factor for Resilience} 
State estimation in challenging environments demands not only sensor redundancy but also computational efficiency.  However, most existing odometry frameworks rely on rigid multi-modal fusion strategies that prioritize robustness by incorporating additional sensors, yet still fail to generalize across diverse degradation scenarios~\cite{prorok2021beyond, ebadi2023present}.

\changed{This limitation arises from two common assumptions: measurement noise follows a  Gaussian distribution, and averaging redundant measurements yields accurate estimates. However, in practice, these assumptions break down in degraded environments such as smoke or darkness, where sensor signals deviate from expected distributions and corrupted measurements dominate certain directions of the state space. As a result, state estimation becomes unreliable, and optimization grows computationally expensive due to the increasing difficulty of distinguishing inliers from outliers. }
This highlights a critical need for foundational research on resilient systems.

To address this, we propose a hierarchical adaptation framework for resilient odometry. Our approach dynamically reconfigures the system based on degradation severity, starting with lightweight adaptations and escalating to more complex ones when needed. This multi-level strategy enables reliable state estimation across diverse degradation types, including visual, geometric, mixed, and full sensor failures (see Fig.~\ref{fig:comprehensive_result}).

\paragraph{Low-cost IMUs are More Useful than Expected}

Historically, IMUs have been treated as auxiliary or secondary sensors in odometry pipelines, typically limited to pre-integration~\cite{forster2015imu}. However, we argue that IMUs hold far greater potential and can serve as independent sources for state estimation, contributing as meaningfully as LiDAR and visual sensors.

We introduce a reciprocal fusion strategy in which learning-based IMU odometry and traditional model-based sensor fusion continuously learn from each other to enhance overall performance. Our approach enables the IMU model to adapt and improve using feedback from the sensor fusion pipeline, while simultaneously providing motion priors and acting as a reliable fallback solution under degraded conditions. This bidirectional adaptation leads to improved accuracy and robustness across diverse environments.

\paragraph{Improving Sensor Calibration and Time Synchronization} 

Currently, the performance of our system depends heavily on accurate calibration and precise time synchronization. \changed{Implementing online calibration and time-sync techniques ~\cite {lv2022observability} could eliminate the need for manual parameter tuning. A promising direction is to leverage continuous-time batch optimization~\cite{furgale2012continuous} combined with observability-aware calibration~\cite{lv2022observability}, which actively selects informative motion segments to align sensor trajectories. This would enable the joint estimation of intrinsic parameters, extrinsic transformations, and temporal offsets in a unified framework.}

\paragraph{Improving Generalizability of the IMU model} The learning-based IMU model requires faster adaptation to new robots and environments. Despite generalizing well across platforms, it struggles with unseen domains due to distribution gaps between training and testing data. Incorporating both real-world and simulated IMU data could reduce this gap and improve generalization.

In conclusion, the proposed hierarchical adaptation system offers robust, adaptable, and efficient state estimation in degraded environments. Although challenges remain, such as time synchronization, calibration, and improvements to the generalization of learning IMU odometry, this work lays a strong foundation for advancing resilient odometry under real-world conditions.

\begin{figure*}[!t]
    \centering
    \includegraphics[width=\textwidth,height=0.78\textheight,keepaspectratio]{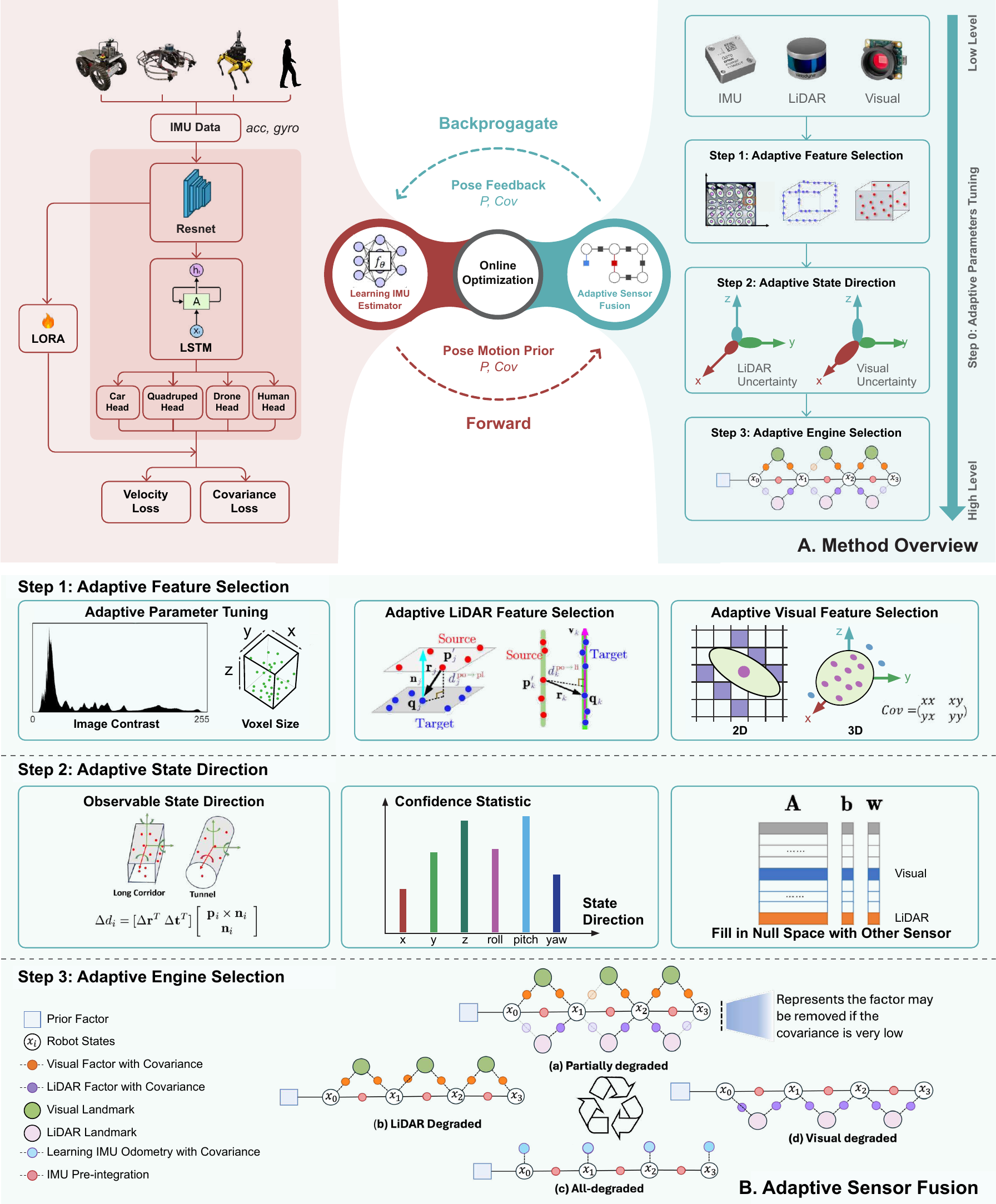}
    \caption{\textbf{System Overview.} \changed{\textbf{(A)} A hierarchical adaptation framework that integrates adaptive sensor fusion with a learning-based IMU odometry. The sensor fusion module provides free pose supervision to guide the learning process of the IMU odometry, helping to reduce drift and capture motion patterns. In turn, the IMU odometry offers motion priors and serves as a fail-safe solution. \textbf{(B)} Illustration of adaptive feature selection, adaptive state direction estimation, and adaptive engine selection.}}
    \label{fig:method}
\end{figure*}

\section*{Materials and Methods}

Before outlining our methodology, we first clarified the key technical terms used in this paper. We proposed hierarchical multi-level mechanisms to address environmental degradation, ranging from mild to extreme. These mechanisms included adaptive feature selection, adaptive state direction, adaptive engine selection, and learning-based inertial odometry. For clarity, we grouped adaptive feature selection, state direction, and engine selection under the term adaptive sensor fusion, as they were based on classic factor graph optimization. We treated learning-based inertial odometry separately, as it relied on a deep learning network. Together, these components formed what we defined as a hierarchical adaptation framework.

\subsection*{Problem Definition}

\paragraph{State Estimation Definition}

\changed{Following the SLAM formulation~\cite{cadena2016past,thrun2002probabilistic}, the general goal of state estimation is to infer the agent's state \( x_t \), and optionally the map \( m_t \), from a sequence of control inputs \( u_{1:t} \) and observations \( o_{1:t} \), via the posterior:}
\begin{equation}
P(x_t, m_t \mid o_{1:t}, u_{1:t}).    
\end{equation}

In contrast to SLAM, the odometry problem focuses solely on estimating the agent's trajectory over time, without building a persistent map. The objective is to estimate the current or next pose \( x_{t+1} \) given a sequence of past observations \( o_{1:t} \) and control inputs \( u_{1:t} \). Odometry aims to compute:
\begin{equation}
P(x_{t+1} \mid o_{1:t}, u_{1:t}).    
\end{equation}
\changed{
Assuming a probabilistic motion model \( P(x_{t+1} \mid x_t, u_{t+1}) \) and an observation model \( P(o_{t+1} \mid x_{t+1}) \), the belief over the current state is recursively updated using Bayes' rule:}
\begin{equation}
\begin{aligned}
P(x_{t+1} \mid o_{1:t+1}, u_{1:t+1}) &={} \eta \cdot P(o_{t+1} \mid x_{t+1}) \\
&\quad \times \int P(x_{t+1} \mid x_t, u_{t+1}) \\
&\qquad \cdot P(x_t \mid o_{1:t}, u_{1:t}) \, dx_t,
\end{aligned}    
\label{eq:slam}
\end{equation}
\changed{
where \( \eta \) is a normalization constant. Based on the above, odometry problem is incrementally estimating the next pose \( x_{t+1} \)  by given local observations and short-term motion models.}

\paragraph{Robustness Definition} Existing evaluation metrics such as ATE~\cite{Burri25012016} have limitations in evaluating the odometry's robustness in real-world applications. 
ATE primarily focused on the overall trajectory error and can not effectively capture the robustness of the algorithm. To evaluate the robustness, we used the relative pose between frames as robustness metrics \cite{zhao2021super}. Unlike traditional Relative Pose Error (RPE) evaluations, our metric assesses both the local accuracy of the trajectory and its completeness across the entire trajectory. The robustness metric is the Area Under the Curve (AUC) of the F-1 score:
\begin{equation}
F_1(e) = \frac{2P(e<{T})R(e<{T})}{P(e<{T}) + R(e<{T})},
\end{equation}
where the precision $P$ quantifies the precision of the relative pose: how closely the estimated relative pose matches the ground truth point, and the recall rate $R$ measures the completeness, to what extent the estimated relative pose covers all ground truth points. Specifically, an estimated error $e$ is regarded as an inlier,  if it is smaller than a threshold ${T}$. \changed{To evaluate robustness across different tolerance levels, we apply an exponential mapping} \changed{ \( \exp(-10T) \) to normalize the threshold range to \([0, 1]\). We then compute the area under the F1 score curve (AUC) by sweeping across this range of thresholds. This provides a comprehensive, threshold-independent measure of overall robustness.}

\paragraph{Position Robustness $R_{p}$ \& Rotation Robustness $R_{r}$} The AUC of F-1 score can be defined using both linear velocity and angular velocity, which can reflect the robustness of position and rotation estimation, respectively. Specifically, we define $R_p = \rm{AUC}(F_1(v_e))$ and $R_{r} = \mathrm{AUC}(F_1(\omega_e))$, where $v_e$ and $\omega_e$ are the estimated error of relative position and orientation between frames, respectively.

\subsection*{Overview of Approach}

The core of our method was a hierarchical adaptation framework that integrated adaptive sensor fusion with learning-based IMU odometry in a self-supervised manner, as shown in Fig.~\ref{fig:method}A. This framework was structured as a bilevel optimization \cite{wang2025imperative,fu2024islam}, where adaptive sensor fusion is in the lower-level (a ``teacher''), providing optimization-based constraints for the IMU odometry in the upper-level (a ``student'').
When vision or LiDAR functions well, IMU odometry receives feedback to learn robot motion. When vision or LiDAR degrades, the IMU module takes over to maintain reliable estimation.
Together, this cooperative symbiosis significantly enhances the
robustness, resilience, and adaptability of odometry systems in challenging environments. For details, please see the System Overview section in the Supplementary.

\FloatBarrier

\subsection*{Adaptive Sensor Fusion} 
We proposed adaptive sensor fusion that hierarchically selects information from parameters, features, states, and engines across multiple levels. It included adaptive feature selection, adaptive state selection, adaptive engine selection, and adaptive parameter tuning.

\paragraph{Adaptive Feature Selection~}
As shown in Fig.~\ref{fig:method}B-Step 1, not all features (visual or geometric) contributed equally to state estimation. Thus, selecting the most informative ones is essential. 
For visual features, we used the Hessian matrix of KLT tracking\cite{hwangbo2009inertial} and analyzed its covariance $\boldsymbol{\Sigma}_{\mathrm{SE(2)}}$ using small patch sizes \cite{muhle2022probabilistic} from intensity residuals. Given the estimated $\boldsymbol{R}_\theta \in \mathrm{SO}(2)$, the rotated covariance for each 2D feature is:
\begin{align}
    \boldsymbol{\Sigma}_{2 \mathrm{D}, \theta}=\boldsymbol{R}_\theta \boldsymbol{\Sigma}_{\mathrm{SE(2)}} \boldsymbol{R}_\theta^{\top}. 
\end{align}
For LiDAR, we evaluated the quality of correspondences based on the fit of neighboring points to point, line, or plane distributions \cite{demantke2011dimensionality}:
\begin{equation}
 w_i^{\mathrm{po} \rightarrow \mathrm{li}}= \frac{\sigma_1 - \sigma_2}{\mu}, \quad  w_i^{\mathrm{po} \rightarrow \mathrm{pl}} = \frac{\sigma_2 - \sigma_3}{\mu}, \quad  w_i^{\mathrm{po} \rightarrow \mathrm{po}} = \frac{\sigma_3}{\mu},    
\end{equation}
where \( \sigma_i \) are the eigenvalues of the Principal Component Analysis (PCA) used for point-to-point, point-to-line, point-to-plane distance error metric\cite{zhao2021super}. The normalization coefficient \( \mu = \sum_{i=1}^3 \sigma_i \) is used and
\( W_l = \{ w_i^{\mathrm{po} \rightarrow \mathrm{po}}, w_i^{\mathrm{po} \rightarrow \mathrm{li}}, w_i^{\mathrm{po} \rightarrow \mathrm{pl}} \} \in [0, 1] \) describe the linearity (\( w_i^{\mathrm{po} \rightarrow \mathrm{li}} \)), planarity (\( w_i^{\mathrm{po} \rightarrow \mathrm{pl}} \)), and scatter (\( w_i^{\mathrm{po} \rightarrow \mathrm{po}} \)) of the features. This approach enables the selection of the meaningful features while reducing the computational load.

\paragraph{Adaptive State Direction~} Various forms of degradation hinder the robustness of existing odometry systems \cite{xu2021fast}\cite{LVI-SAM}. However, \changed{existing approaches} require prior knowledge of the environment to determine well-conditioned state estimation and cannot adapt very well to changing environments\cite{degeneracy,zhen2019estimating,ding2021degeneration,hinduja2019degeneracy}. Moreover, some methods perform degeneracy analysis \cite{ebadi2021dare} \changed{or estimate a} robust kernel function \cite{chebrolu2021adaptive} during or after the optimization. \changed{However, this may} be too late to predict potential failures \cite{degeneracy} and insufficient data association already causes unreliable optimization.

To address potential failures, we performed a degeneracy analysis by evaluating the observability of LiDAR odometry at the front-end (Fig.~\ref{fig:method}C-Step 2). We focused on LiDAR rather than vision, as it offers more reliable observability and uncertainty estimates for each state direction. Once LiDAR quantifies directional uncertainty, visual weights can be adjusted accordingly. Specifically, we predicted ICP alignment risk and computed the observability \(\mathbf{Obs}_i\) for each point along each state direction. Each correspondence was then labeled based on the state it primarily influences. Due to space constraints, please refer to the Supplementary Materials for details on alignment risk prediction and observability for each point-to-plane constraint.

To evaluate observability over a scan, we developed a statistical method that analyzes label distributions across state directions. Let $\mathcal{O} = \{x, y, z, \text{roll}, \text{pitch}, \text{yaw}\}$ be the set of observable dimensions in our state space. For each dimension $i \in \mathcal{O}$, we defined $N_i$ as the count of observability labels in that dimension.
The total observability count was given by $N_{\text{total}} = \sum_{i \in \mathcal{O}} N_i$. We defined the normalized confidence metrics as:
\begin{align}
\boldsymbol{\gamma}_{\text{trans}} &= {|\mathcal{O}|} \begin{bmatrix}
\frac{N_x}{N_{\text{total}}}, & 
\frac{N_y}{N_{\text{total}}}, & 
\frac{N_z}{N_{\text{total}}}
\end{bmatrix}^T \\
\boldsymbol{\gamma}_{\text{rot}} &= {|\mathcal{O}|} \begin{bmatrix}
\frac{N_{\text{roll}}}{N_{\text{total}}}, & 
\frac{N_{\text{pitch}}}{N_{\text{total}}}, & 
\frac{N_{\text{yaw}}}{N_{\text{total}}}
\end{bmatrix}^T
\end{align}
Here, the factor ${|\mathcal{O}|} = 6$ scales the metrics to $[0,1]$ under uniform distribution across all dimensions. 
These confidence metrics construct a diagonal $6 \times6$ covariance matrix:
\begin{align}
\boldsymbol{\Sigma}_{cov} = \text{diag}(\boldsymbol{\gamma}_{\text{trans}}, \boldsymbol{\gamma}_{\text{rot}}).
\label{eq:cov}
\end{align}
It's important to note that the Eq.~\ref{eq:cov} assesses the ratio of observability labels in
a specific direction to the total number of labels. This relative approach is based on the hypothesis that in well-structured environments, observability should be uniformly distributed across all state directions.

By analyzing label distributions across scan correspondences, our approach detects weakly observed directions needing reinforcement. This allows the system to fuse sensor information more effectively, especially along those vulnerable directions.

\paragraph{Adaptive Engine Selection~} In real-world settings, sensors often operate under persistent degradation (e.g., darkness). To reduce computation and filter faulty data, we proposed an adaptive factor selection strategy that dynamically reconfigured the factor graph based on sensor quality (Fig.~\ref{fig:method}B-Step 3). Factor graph optimization should be dynamically reconfigurable based on degradation type. By evaluating factor quality, the system decided whether to include or exclude a sensor's data.

For example, during prolonged visual degradation, where the visual sensor's contribution dropped below 10\% for 2--4 seconds (such as in darkness), the system excluded visual factors and relied on LiDAR. If LiDAR became unreliable due to loss of observability, its factors were also rejected. In mixed degradation scenarios, the system selected the most reliable sensor data. When both LiDAR and vision failed, the system fell back to learning-based IMU odometry.

We adopted a factor graph to represent all residuals, but actively selected them in optimization based on degeneracy. Using prior observability analysis, the covariance matrix $\mathbf{cov}$ guided the inclusion of only well-constrained factors.
\begin{equation}
\begin{aligned}
\min_{\mathbf{T}_{i+1}} \bigg\{ 
& \sum_{p \in \mathbb{F}_i} \left\| e_i^{po \to po,li,pl} \right\|_{\mathbf{cov_L}}^2 
+ \sum_{(i, i+1) \in \mathcal{B}} \left\| e_{io}^{prior} \right\|_{\mathbf{cov_{io}}}^2 \\
& + \sum_{(i, i+1) \in \mathcal{B}} \left\| e_{vo}^{prior} \right\|_{\mathbf{cov_{vo}}}^2 
\bigg\}
\end{aligned}
\end{equation}
In this context, \( e^{prior}_{io} \) represented the relative pose factor provided by learning-based IMU odometry, weighted by the proposed \( \mathbf{cov_{io}} \). Similarly, \( e^{prior}_{vo} \) was the relative pose factor from visual odometry, weighted by \( \mathbf{cov_{vo}} \). The term \( e_i^{po \to po,li,pl} \) indicated point-to-point (line, distance) residuals between two keyframes with covariance \( \mathbf{cov_{L}} \)\cite{zhao2021super}. For more on the relative pose factor, please refer to the Supplementary. This adaptability ensured our sensor fusion remains flexible and resilient, dynamically optimizing performance under varying conditions.

\paragraph{Adaptive Parameter Tuning~} As shown in Fig.~\ref{fig:method}B-Step 1, our method used self-tuning strategies to ensure adaptability across diverse platforms and environments\cite{ramezani2022wildcat, lim2023adalio}. For visual odometry, we adjusted image contrast to improve feature detection~\cite{bergmann2017online}. LiDAR odometry, however, may diverge in narrow spaces (e.g., stairwells or corridors) due to parameters optimized for open environments. To address this, the algorithm dynamically adjusted the weighting between visual and LiDAR odometry to adapt to both confined and open areas. For details, please see the Adaptive State Direction in Multi-floor Environments Experiment in the Supplementary.

Other parameters, such as voxel size, were adapted to the environment scale. The algorithm also used intensity values to filter airborne obscurants~\cite{best2022resilient} and gravity vectors to estimate initial rotational alignment between LiDAR and IMU~\cite{nemiroff2023joint}.

\subsection*{Learning-based Inertial Odometry} 

We introduced a learning-based inertial odometry system designed to capture motion patterns across various robots. The structure was shown in Fig.~\ref{fig:method}A. It consisted of two key components: a pre-trained IMU model trained on large datasets, and an adapter network for unseen environments.

To enable generalization and few-shot capability, the pre-trained model was trained on diverse datasets, including wheeled, legged, aerial robots, and handheld devices. It includes five modules: IMU data preprocessing, data augmentation, a heterogeneous shared backbone, a multi-head design, and a loss function (Fig.~\ref{fig:method}A). We first describe the pre-trained IMU model, then its adapter network.

\paragraph{Data Preprocessing}

Our training dataset included over 100 hours of real-world IMU data from diverse platforms-wheeled robots, drones, legged robots, and human-held devices---sourced from SubT-MRS \cite{zhao2024subt}, TartanDrive \cite{sivaprakasam2024tartandrive}, IDOL \cite{sun2021idol}, Blackbird \cite{antonini2018blackbird}, and UZH \cite{Delmerico19icra}. These datasets provide both raw IMU measurements and ground-truth trajectories across eight platforms with varying dynamics.
To ensure consistency, we unified all IMU axes to a common frame (X-forward, Y-left, Z-up), resampled all data to 200 Hz via interpolation. This preprocessing reduced variability across sources and ensured uniform input quality during training and inference.

\paragraph{Data Augmentation}
Inspired by the work of Cao et al. \cite{cao2022rio}, we rotated the IMU data around the X, Y, and Z axes by a specified degree. Simultaneously, the corresponding ground truth trajectory was also rotated by the same angle. A pre-trained neural network for the IMU was trained on both the original and rotated datasets, as illustrated in Fig.~\ref{fig:method}B. This approach significantly enlarges the IMU training dataset and enhances the model's generalization capability.  

\paragraph{Heterogeneous Shared Backbone}

After preprocessing and augmentation, IMU data were passed through a heterogeneous shared backbone designed to generalize across diverse robot platforms. This backbone maps data into a unified high-dimensional latent space that captures common motion knowledge, enabling efficient adaptation to new platforms with minimal additional training.

As shown in Fig.~\ref{fig:method}A, the network consisted of three key components: a 1D ResNet backbone, LSTM layers, and a multi-head output. The ResNet extracted motion features from the IMU data; the LSTM modeled temporal dependencies; and the multi-head layer predicted local velocity and its covariance, completing the motion estimation pipeline.

\paragraph{Multi-head}

Building on a shared heterogeneous backbone, we introduced a multi-head architecture to map latent features to velocity estimates, with each head tailored to a specific robot type (Fig.~\ref{fig:method}B). This architecture offered two main advantages: (i) it enabled learning of diverse motion patterns in parallel, enhancing adaptability across different robot types; and (ii) it stabilized training by avoiding conflicting learning objectives. The mathematical formulation is as follows:
\begin{equation}
\begin{aligned}
(\hat{v}, \hat{u}) &= f\left(\left( ^B \mathbf{a}_{n-N}, ^B \boldsymbol{\omega}_{n-N} \right), \dots, \left( ^B \mathbf{a}_n, ^B \boldsymbol{\omega}_n \right), \mathbf{h}_{n-N} \right) \\
^B \mathbf{a}_n &= ^B_W\mathbf{R}_n (^W_B \mathbf{R}_n \left( \mathbf{a} - \mathbf{b}_a \right) - ^W \mathbf{g} )\\
^B \boldsymbol{\omega}_n &= \boldsymbol
{\omega} - \mathbf{b}_g.
\end{aligned}
\end{equation}
Here, \( f(\cdot) \) represents the function defined by the neural network that processes inputs from the IMU sensor. The network receives as input the acceleration \(\mathbf{a}\), angular velocity \(\boldsymbol{\omega}\), and the hidden state \(\mathbf{h}_{n-N}\) produced by the LSTM at the previous time step. For each IMU measurement, we removed the gravity vector \( ^W \mathbf{g} = [0, 0, 9.8] \) as well as the acceleration bias \(\mathbf{b}_a\) and gyro bias \(\mathbf{b}_g\). At each time step, the network predicts the current motion based on the hidden state \(\mathbf{h}_{n-N}\) and a local window of \(N\) samples of acceleration and angular velocity in the inertial frame \(B\). The output of the network consists of the estimated relative velocity \(\hat{{v}}\) and its associated uncertainties \(\hat{{u}}\).

Unlike existing learning-based IMU odometry methods~\cite{liu2020tlio, IMO}, which process IMU data in world coordinates ($W$), our method operates entirely in the body frame ($B$). \changed{Specifically, we first transformed all IMU data into a unified body frame (Z-up, Y-left, X-forward). We then directly used raw acceleration ($a$) and angular velocity ($\omega$) in the body frame as input to the network and regressed velocity in the body frame. Because both the input and loss functions were defined in the body frame, our method reduced the risk of overfitting to specific trajectories in the world frame and exhibited improved generalization. }

\paragraph{Loss Function}
Our network employed relative loss functions, as illustrated in Fig.~\ref{fig:method}~B. The relative loss function helped the network capture motion dynamics within a single window, improving the smoothness of the predicted odometry. To optimize the relative loss, we utilized the Mean Squared Error (MSE), defined as follows:
\begin{align}
L_{RL}^{MSE}(\mathbf{v}, \hat{\mathbf{v}}) = \frac{1}{n} \sum_{i=1}^{n} \left(\sum_{j=m}^{m+M} \left( v_{j \rightarrow j+1} - \hat{v}_{j \rightarrow j+1} \right)\right) 
\end{align}
We defined the loss in the body frame. Specifically, \(\hat{v}_{j \rightarrow j+1}\) represented the 3D velocity output of the network for the \(j\)-th window in the body frame, while \(v_{j \rightarrow j+1}\) denoted the corresponding ground truth in the same frame. Variable \(n\) is the batch size during training, \(m\) is the starting window of the sequence, and \(M\) is the total number of LSTM windows.

\paragraph{\textbf{Covariance:}} The covariance estimation follows the methodology outlined in \cite{liu2020tlio}. The network outputs a three-dimensional vector, where each element represents the logarithm of a diagonal element of the covariance matrix $\Sigma $. For more implementation details on learning imu odometry, please refer to our Supplementary. 
\begin{align}
L^{\text{NLL}} = \frac{1}{2} (v - \hat{v})^T \Sigma^{-1} (v - \hat{v}) + \frac{1}{2} \ln \left| \Sigma \right|
\end{align}

\paragraph{Adaptor Network}
Although the pre-trained IMU model was designed to generalize across platforms, it cannot guarantee ``few-shot'' capability for every robot or motion pattern. To address this limitation, we introduced a  lightweight Low-Rank Adaptation (LoRA) module~\cite{hu2022lora} that enabled the model to acquire new knowledge and improve continuously as it processed additional data. This adapter was integrated into the pre-trained IMU model (Fig.~\ref{fig:method}B).
To update each weight matrix \( W_0 \in \mathbb{R}^{d \times k} \) of the pre-trained IMU model, we applied a low-rank decomposition: 
\begin{align}
W_0 + \Delta W = W_0 + BA,
\end{align}
where \( B \in \mathbb{R}^{d \times r} \) and \( A \in \mathbb{R}^{r \times k} \), and $r$ is the rank and \( r \ll \min(d, k) \). During adaptation, \( W_0 \) remained fixed, while \( A \) and \( B \) are trainable parameters. Both \( W_0 \) and \( \Delta W = BA \) process the input, and their outputs are summed, resulting in:
\begin{align}
h = W_0 x + \Delta W x = W_0 x + BA x,  
\end{align}
where \( A \) is initialized by a random Gaussian distribution and \( B \) starts at zero. 
By freezing the original model parameters, our adaptive sensor fusion framework transferred knowledge solely to the LoRA adapter, preventing catastrophic forgetting. The adapter's small parameter set enabled rapid assimilation of new information for real-time adaptation. 
For more details, please refer to our experiments on catastrophic forgetting in the Supplementary Material.

This approach ensured flexible adaptation to diverse robotic systems while preserving the pre-trained model's generalization ability. It enabled efficient incorporation of new knowledge and system-specific optimization, without compromising the core capabilities.

\subsection*{Statistic Analysis}
Statistical analysis was performed in Python using the NumPy library to compute the Absolute Trajectory Error (ATE)~\cite{sturm2012benchmark} and robustness metrics~\cite{Zhao2024CVPR}. Evaluations in Fig.~\ref{fig:drop_sensor} and Tables~\ref{tab:AccLidar}--\ref{tab:RobustLidar} were conducted on the SubT-MRS dataset~\cite{Zhao2024CVPR}, which contains eight sequences categorized into geometric and mixed degradation groups. Each algorithm was evaluated across multiple runs per sequence using our ICCV 2023 SLAM Challenge submission system, and results from competing methods were provided by the original authors. For fair comparison, all odometry trajectories were upsampled to 200 Hz. The mean ATE quantified trajectory accuracy, and the robustness metrics jointly assessed accuracy and completeness, providing an integrated measure of system resilience. Sequences where all baseline algorithms failed, e.g., dense smoke in Fig.~\ref{fig:comprehensive_result} and \ref{fig:offroad_smoke}, were excluded from the averaging and omitted from Tables~\ref{tab:AccLidar}--\ref{tab:RobustLidar}.

\section*{Acknowledgments}
We thank Yuheng Qiu, Michale Kaess, Sudharshan Suresh, and Shubham Tulsiani for their valuable suggestions for the manuscript. 
We sincerely appreciate the work of Honghao Zhu, Rushan Jiang, Haoxiang Sun, Tianhao Wu, Yuanjun Gao, Damanpreet Singh, Lucas Nogueira, and Guofei Chen for their help in real-world experiments.\\
% Furthermore, we are truly grateful for Dr. Yonatan Bisk's suggestions regarding the design of the experiment. \\
\textbf{Funding:} U.S. Army Research Lab W911NF-23-S-0001, U.S. Army Research Lab
W911NF2120152, U.S. Army Research Lab W911NF2420125, U.S. Army Research Lab
W911NF-17-S-0003. \\
\textbf{Author Contributions:} \\
Conceptualization: Shibo Zhao, Wenshan Wang, Chen Wang, Ji Zhang, Sebastian Scherer \\
Methodology: Shibo Zhao, Sifan Zhou, Wenshan Wang, Chen Wang  \\
Investigation: Shibo Zhao, Sifan Zhou, Yuchen Zhang \\
Visualization: Shibo Zhao, Sifan Zhou, Ji Zhang, Chen Wang, Sebastian Scherer \\
Funding Acquisition: Sebastian Scherer, Wenshan Wang \\
Project Administration: Shibo Zhao, Wenshan Wang, Sebastian Scherer \\
Supervision: Wenshan Wang, Sebastian Scherer \\
Writing -- Original Draft: Shibo Zhao, Sifan Zhou \\
Writing -- Review \& Editing: Shibo Zhao, Chen Wang, Wenshan Wang, Ji Zhang, Sebastian Scherer, Yuchen Zhang\\
\textbf{Competing Interests:} The authors declare they have no competing interests. \\
\textbf{Data and materials availability:} All data needed to evaluate the conclusions in the paper are present in the paper or the Supplementary Materials.  The data for this study have been deposited in the database \url{https://zenodo.org/records/17569700}\cite{zhao_2025_17569700} \\

\subsection*{Supplementary materials}
% Materials and Methods\\
% Supplementary Text\\
% Figs. S1 to S3\\
% Tables S1 to S4\\
% References \textit{(7-\arabic{enumiv})}\\ % automatically fills out the last reference number
% % (filling out the other numbers automatically is possible but fiddly and liable to break)
% Movie S1\\
% Data S1

% \subsection*{Supplementary Materials}
Supplementary Methods. \\
Supplementary Figures S1 to S7. \\
Supplementary Movies S1 to S6.

\clearpage
%%%%%%%%%%%%%%%% START OF SUPPLEMENT %%%%%%%%%%%%%%%

% Figures, tables, equations and pages in the supplement are numbered S1, S2 etc.
\renewcommand{\thefigure}{S\arabic{figure}}
\renewcommand{\thetable}{S\arabic{table}}
\renewcommand{\theequation}{S\arabic{equation}}
\renewcommand{\thepage}{S\arabic{page}}
\renewcommand{\theHfigure}{supp.\arabic{figure}}
\renewcommand{\theHtable}{supp.\arabic{table}}
\renewcommand{\theHequation}{supp.\arabic{equation}}
\setcounter{figure}{0}
\setcounter{table}{0}
\setcounter{equation}{0}
\setcounter{page}{1} % not 0 as \newpage already started a supplementary page
% References continue the numbering from the main text.

%%%%%%%%%%%%%%%% SUPPLEMENT TITLE PAGE %%%%%%%%%%%%%%%

\begin{center}
\section*{Supplementary Materials for\\ \scititle}

% Author list for the supplement
% Indicate the corresponding authors, but do NOT include institutions here
% It would be nice if the template auto-generated this, but doing so is complicated...
% First~Author$^{\ast\dagger}$,
% A.~Scientist$^\dagger$,
% Someone~E.~Else\\ % we're not in a \author{} environment this time, so use \\ for a new line
Shibo Zhao$^{1\ast}$,
	Sifan Zhou$^{1}$,
    Yuchen Zhang$^{1}$,
    Ji Zhang$^{1}$,
    Chen Wang$^{2}$, \\
    Wenshan Wang$^{1\dagger}$,
	Sebastian Scherer$^{1\dagger}$ \\

    \small$^{1}$Carnegie Mellon University, USA.\\
    \small$^{2}$University at Buffalo, USA.\\
    \small$^\ast$Corresponding author. Email: shiboz@andrew.cmu.edu\\
	% Joint contributions can be indicated like this
\small$^\dagger$Equally Advising.
\end{center}

% Fill out the numbers for each type of supplementary material,
% and delete any lines that aren't applicable.
% These are just example numbers that don't match the rest of this template.
\subsubsection*{This PDF file includes:}
% Materials and Methods\\
% Supplementary Text\\
% Figures S1 to S3\\
% Tables S1 to S4\\
% Captions for Movies S1 to S2\\
% Captions for Data S1 to S2

\begin{figure*}[!t]
   \centering
    \includegraphics[width=1.0\textwidth,height=0.84\textheight,keepaspectratio]{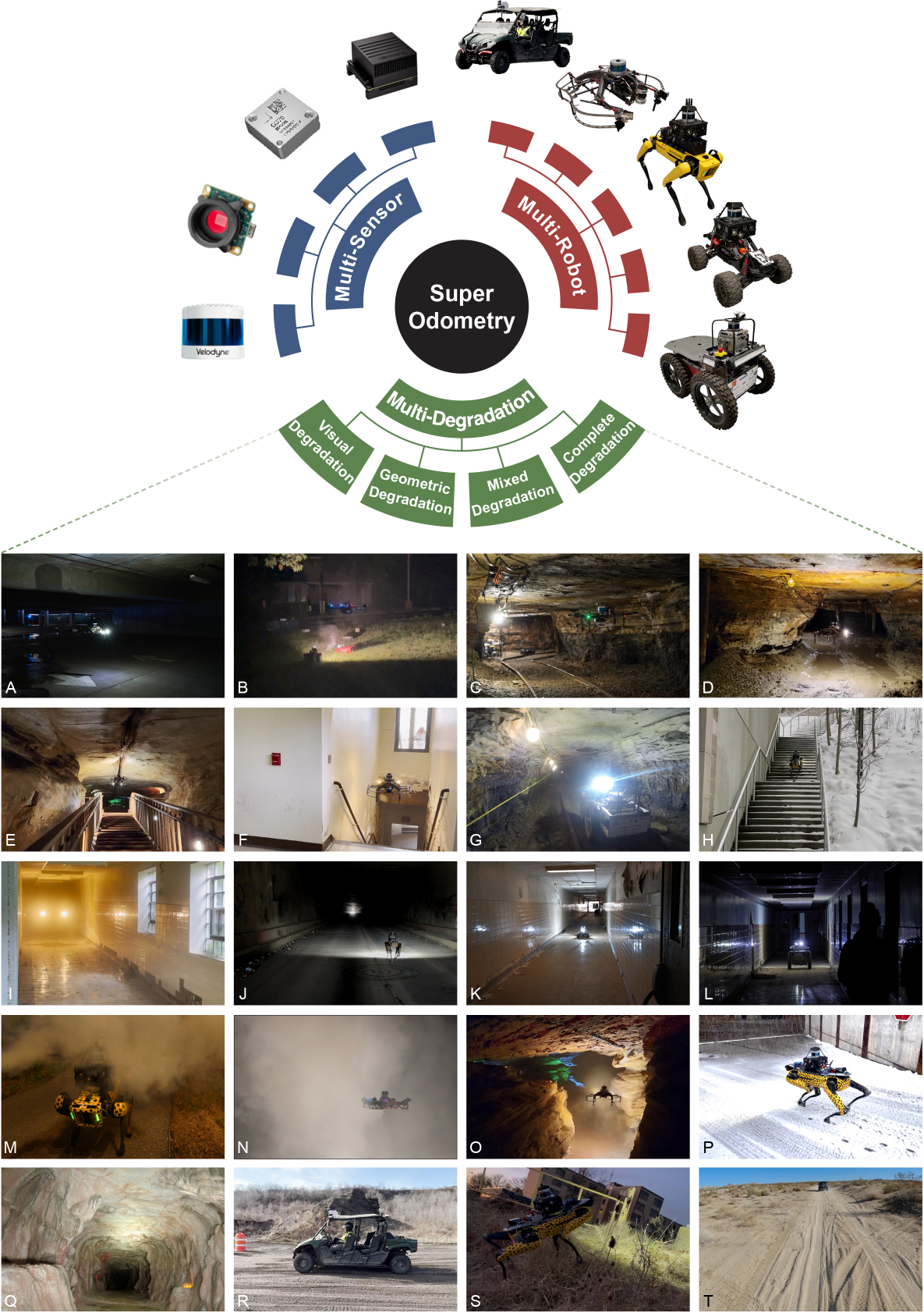}
    \caption{\textbf{Robust odometry across diverse conditions with various sensors and robots.}  We tested our method in various scenarios with visual degradation (A-D), geometric degradation (E-H), mixed degradation(I-L), complete degradation (M-P) and other scenarios such as caves, off-road, steep slopes, and featureless deserts (Q-T).} 
    \label{fig:various_scenes}
\end{figure*}

System Overview \\
Supplementary Methods.\\
\fref{fig:various_scenes} Robust odometry across diverse conditions with various sensors and robots. \\
\fref{fig:subt_final} Large-scale mapping results from the DARPA subterranean challenge. \\
\fref{fig:multi_floor} Illustration of adaptive state selection and self-weight parameter tuning in a multi-floor environment. \\
\fref{fig:offroad} Aggressive motion in off-Road environments. \\
\fref{fig:subt_drone} Unexpected Collision from a Spinning Drone. \\
\fref{fig:snow} Mapping Results in Snow Environments. \\
\fref{fig:no_forget} Non-forgetting Characteristics of Our Domain Adaption. \\
\tref{tab:network_setting} Learning-based IMU Odometry Network Configuration.
\subsubsection*{Other Supplementary Materials for this manuscript:}
Movie S1 Method Summary Introduction. \\
Movie S2 Evaluation of 13 Types of Degradation in a Single Run.\\
Movie S3 Illustration on Adaptive State Direction in a Long Corridor Environments. \\
Movie S4 Performance of IMU Pre-trained Model.  \\
Movie S5 Robust Performance in Smoke Scenario.  \\
Movie S6 Robust Odometry Across Diverse Conditions with Various Sensors and Robots. \\
% Movie S7 Multi-Robot State Estimation in Challenging Environments. \\
% Movie S8 Robust Odometry Performance for Exploration Tasks Using Onboard Device.

% \subsubsection*{Other Supplementary Materials for this manuscript:}
% % Movies S1 to S2\\
% % Data S1 to S2
% Movies S1 to S6

\newpage

\subsection*{System Overview}
Super Odometry integrates an adaptive sensor fusion strategy with a learning-based IMU odometry in a self-supervised manner. The adaptive sensor fusion framework assesses degeneracy from low to high levels and provides supervision signals to the learning-based inertial state estimator. In return, the inertial state estimator learns the robot motion dynamics and offers motion priors to the sensor fusion framework. In addition, it allows the system to run without failure for up to a few minutes under extreme degradation, where all exteroceptive sensors fail.   The following paragraphs provide an overview of the adaptive sensor fusion, including adaptive feature selection, adaptive state direction, adaptive engine selection, and learning-based inertial odometry. 

\subsubsection*{Adaptive Sensor Fusion}  
Handling multiple degradations is a major challenge for odometry systems. 
Although several strategies have been proposed to alleviate this issue, e.g., degeneracy mitigation \cite{tagliabue2021lion,tuna2023x}, uncertainty estimation\cite{talbot2023principled,brossard2020new} and observability analysis \cite{degeneracy,zhen2019estimating,ding2021degeneration,hinduja2019degeneracy,tagliabue2021lion} \changed{and multi-sensor fusion\cite{r3live, shan2021lvi, nubert2025holistic, zhong2021lvio}}, none of these can overcome all types of degraded environments.

To solve this, we introduced adaptive sensor fusion, including adaptive feature selection, adaptive state direction, and adaptive engine selection.

\textit{\textbf{Adaptive Feature Selection:}~}
In visually degraded environments, we observed that not all visual features contribute equally to state estimation (Fig.~\ref{fig:concept} B). Therefore, we prioritized the most valuable features by analyzing the uncertainty associated with each feature~\cite{muhle2022probabilistic}.

\textit{\textbf{Adaptive State Direction:} ~}
In geometrical degradation, some directions, such as along a long and windowless corridor, cannot be well constrained by the LiDAR odometry shown in Fig.~\ref{fig:concept} B.
Our method can identify these poorly constrained directions and actively incorporate pose priors from other odometry sources to enhance robustness.

\textit{\textbf{Adaptive Engine Selection:} ~}
To address mixed degradation from visual and geometric sources, we further employed factor or engine-level uncertainty analysis to actively switch auxiliary sensors based on degeneracy type (Fig.~\ref{fig:concept} B).

\subsubsection*{Heterogeneous Learning-based Inertial Odometry} 

The emergence of deep learning offers new possibilities for extracting information from IMU data and developing motion models for robots. 
Several studies such as IONet \cite{chen2018ionet}, TLIO \cite{liu2020tlio}, RIDI \cite{yan2018ridi}, and RoNIN \cite{chen2021rnin} have demonstrated that learning-based IMU estimators are promising in position estimation and can be used in completely degraded environments, such as smoke as shown in Fig.~\ref{fig:concept} B. However, most of these methods are trained on one dataset for one specific platform. They do not generalize across datasets or platforms~\cite{buchanan2022deep}. To solve this, we made two efforts. 

\textit{\textbf{Pre-trained IMU Model on Large Datasets:}}
we proposed a \textit{general} IMU model, featuring a shared backbone that enables cross-platform motion estimation. It
is trained on large, high-quality, and diverse IMU data \cite{zhao2024subt} \cite{sivaprakasam2024tartandrive} comprising hundreds of hours of data from various platforms, including drones, quadrupeds, cars, and humans, in a variety of scenarios, including cave exploration, drone racing, and indoor navigation. This broad training enabled our \textit{general} model to outperform any specific models~\cite{chen2018ionet,liu2020learning,chen2021rnin,yan2018ridi} and to be applied to various robot platforms with different motion patterns.

\textit{\textbf{Online Adaptation to Unseen Datasets:}} Although our model was trained on a large dataset, there are always out-of-distribution motion patterns that are not covered. 

Furthermore, collecting sufficient data to fine-tune the pre-trained IMU network during deployment was not always feasible. This required the network to adapt itself \textit{seamlessly} to changing environments and platforms. To address this challenge, we proposed an online adaptation scheme for our pre-trained IMU model, enabling it to adjust to novel environments continuously in a self-supervised manner.
Unlike conventional learning approaches where the pre-trained model remains static during inference, our method eliminates the boundary between the training and testing phases --- we learn as we operate. To enable fast adaptation, we integrated Low-Rank Adaptation (LoRA) \cite{hu2022lora} into our online adaptation process, significantly boosting real-time performance. Our inertial odometry model achieved 200 FPS with real-time refinement.

To the best of our knowledge, this was the first IMU-based estimator that demonstrates such a high level of generalization across diverse robot platforms, fast adaptation to unseen environments, and real-time predictions with minimal drift.

\subsubsection*{A Unified Solution for Various Applications}
In this section, we briefly introduced the different sensor configurations, experiment environments, and robot platforms tested with our algorithm.

\textit{\textbf{Multi-sensor Integration:}} An ideal odometry solution should be scalable and adaptable to various sensor types and configurations. Our pipeline adopted factor graph sensor fusion~\cite{dellaert2017factor}, which offered a simpler and more flexible pipeline to integrate various sensors, including visual, thermal, LiDAR, IMU, and GPS. To push the limits of our approach, we deliberately exclude GPS and thermal data in all experiments to increase the complexity of experiments. Additionally, we avoided using any backend optimization techniques for all experiments.

\textit{\textbf{Multi-robot Adaptability:}} A robust odometry system must accommodate varying speeds and agility across different robotic platforms. 
For instance, drones are usually more agile than ground vehicles, while some ground vehicles can reach high speeds.

We tested our method on various platforms, including aerial, wheeled, and legged robots, as well as hand-carried sensors, under various conditions involving aggressive motion. Notably, all experiments were run in real-time on the robots' embedded computing systems.

\textit{\textbf{Range of Degradation:}} 
The proposed odometry system was rigorously tested in perceptually degraded environments and with different types of robots, including aerial, wheeled, and legged platforms. These tests included visually degraded environments (Fig~\ref{fig:various_scenes}A-D), such as textureless indoors, complete darkness, and water puddles, which made visual feature detection and tracking unreliable. Additionally, the system successfully handled geometric degradation (Fig~\ref{fig:various_scenes}E-H), such as steep multi-floor staircases and long tunnels where LiDAR-based systems were prone to drift. It also performed well in mixed degradation (Fig~\ref{fig:various_scenes}I-L), including dark featureless corridors, highly reflective white walls, long textureless tunnels, and structureless passages, which challenge both visual and LiDAR-based approaches. Furthermore, our method was tested in completely degraded environments (Fig~\ref{fig:various_scenes}M-P), including the presence of obscurants like dense smoke, dust, and heavy snow. Beyond these, it was evaluated in several challenging terrains, including self-similar caves, fast off-road, steep bush-covered slopes, and featureless deserts (Fig~\ref{fig:various_scenes}Q-T).

Remarkably, our method consistently performed without any failure, demonstrating exceptional robustness and reliability in all tested conditions, which accumulated 200 kilometers and 800 operational hours. For more details on results, please refer to Supplementary Movie S6.

\subsection*{Supplementary Method}

\subsubsection*{Predicting Alignment Risk in Point Cloud Registration}

Detecting alignment risks before odometry failure is critical for achieving high accuracy and robustness, particularly in extremely challenging environments. This section outlines a method to quantify alignment risk when registering two point clouds, \( P \) and \( Q \), using point-plane correspondences.

Our approach leverages a KD-tree to extract \( k \) point pairs \( (\mathbf{p}_i, \mathbf{q}_i) \), where each point \( \mathbf{q}_i \) is associated with a corresponding normal vector \( \mathbf{n}_i \). The objective is to determine the optimal rigid-body transformation---comprising rotation \( R \) and translation \( t \) to minimize the sum of squared distances between each point \( \mathbf{p}_i \) and the tangent plane of \( Q \) at \( \mathbf{q}_i \).
The alignment error is defined as:

\begin{equation}
E = \sum_{i=1}^{k} \left( (R \mathbf{p}_i + t - \mathbf{q}_i) \cdot \mathbf{n}_i \right)^2
\end{equation}

Expanding this expression, the error can be rewritten as:

\begin{equation}
E = \sum_{i=1}^{k} \left( \left( (\mathbf{p}_i - \mathbf{q}_i) \cdot \mathbf{n}_i \right) + \mathbf{r} \cdot (\mathbf{p}_i \times \mathbf{n}_i) + \mathbf{t} \cdot \mathbf{n}_i \right)^2
\end{equation}

Here, \( \mathbf{r} \) represents the rotation parameterized as a vector. 

Next, we express the point-plane distance transformed by the vector \( [\Delta\mathbf{r}^T \quad \Delta\mathbf{t}^T] \) as:

\begin{equation}
\Delta d_i = 
\begin{bmatrix}
\Delta \mathbf{r}^T & \Delta \mathbf{t}^T
\end{bmatrix}
\begin{bmatrix}
\mathbf{p}_i \times \mathbf{n}_i \\
\mathbf{n}_i
\end{bmatrix}
\end{equation}

This equation reveals that points where \( \mathbf{n}_i \) is orthogonal to \( \Delta \mathbf{t} \), or where \( \mathbf{p}_i \times \mathbf{n}_i \) is orthogonal to \( \Delta \mathbf{r} \), do not contribute to the error \( E \). This insight enables a direct assessment of whether the current motion, represented by \( [\Delta \mathbf{r}^T \quad \Delta \mathbf{t}^T] \), is constrained by analyzing the orthogonality between these vectors.

For each scan, we evaluated the rotational and translational stability of individual correspondences to determine their contribution to the current motion constraints. This analysis results in the construction of the matrix \( C \), which is used to assess alignment quality~\cite{gelfand2003geometrically}:

\begin{equation}
C =
\begin{bmatrix}
\mathbf{p}_1 \times \mathbf{n}_1 & \cdots & \mathbf{p}_k \times \mathbf{n}_k \\
\mathbf{n}_1 & \cdots & \mathbf{n}_k
\end{bmatrix}
\begin{bmatrix}
(\mathbf{p}_1 \times \mathbf{n}_1)^T & \mathbf{n}_1^T \\
\vdots & \vdots \\
(\mathbf{p}_k \times \mathbf{n}_k)^T & \mathbf{n}_k^T
\end{bmatrix}
\end{equation}

This formulation provides a systematic approach to evaluate alignment risk and guide robust motion estimation under challenging conditions.

\subsubsection*{Observability Estimation for Adaptive State Direction}

Despite anticipating potential alignment risks, the observability of current state estimation remains fundamentally uncertain. Existing sensor fusion methods \cite{xu2021fast,wen24liver,zheng24trajlio} employ covariance matrices to quantify state estimation uncertainty, typically assuming Gaussian distribution optimization---even in degenerate scenarios. However, this assumption becomes unreliable when certain states or landmarks become unobservable, potentially introducing numerical instability that compromises the reliability of traditional Gaussian-based marginalization and Schur complement approaches \cite{dellaert2017factor}.

Recognizing that optimization becomes ill-conditioned due to degeneracy, we proposed an alternative strategy: estimating the covariance matrix from the front-end sensor data, rather than deriving it from the optimization framework's Hessian matrix. Leveraging our previous risk prediction insights, we directly analyze the observability of raw sensor measurements at the front-end stage.

For each correspondence point $\mathbf{p}_i$ in the source point cloud $P$, we compute and project six values to the world frame $[\mathbf{X}_w,\mathbf{Y}_w,\mathbf{Z}_w, \mathbf{roll}_w,\mathbf{pitch}_w, \mathbf{yaw}_w]$ using the following formulation:

\begin{equation}
\mathbf{Obs}_i = a^2_{2D}(\mathbf{p}_i)
\begin{bmatrix}
|\mathbf{C}_i \cdot \mathbf{X}_w| \\
|\mathbf{C}_i \cdot \mathbf{Y}_w| \\
|\mathbf{C}_i \cdot \mathbf{Z}_w| \\
|\mathbf{n}_i \cdot \mathbf{X}_w| \\
|\mathbf{n}_i \cdot \mathbf{Y}_w| \\
|\mathbf{n}_i \cdot \mathbf{Z}_w|
\end{bmatrix}
\end{equation}

where $\mathbf{Obs}_i$ is the observability analysis of points $\mathbf{p}_i$ in the world coordinates, $\mathbf{C}_i = T_{init}^{w}\mathbf{p}_i \times \mathbf{n}_i$, $T_{init}^{w}$ is the initial guess of scan registration, $ a^2_{2D}(\mathbf{p}_i)$ is a planner scalar defined by $a_{2D}=(\sigma_2-\sigma_3)/\sigma_1$, and $\sigma_i$ are eigenvalues of Principal Component Analysis (PCA) for normal computation to represent planarity of geometry\cite{demantke2011dimensionality}.

The first three values in $\mathbf{Obs}_i$ quantify the contribution of point $\mathbf{p}_i$ to sensor rotation observability (roll, pitch, yaw). The last three values represent its contribution to the observability of sensor translations. Based on these values, we can assign labels to each correspondence according to which state direction it contributes most. For a more detailed illustration, please refer to our experiments on Adaptive State Direction in Multi-floor Environments in the Supplementary.

\subsubsection*{Relative Pose Prior for Mitigating LiDAR Degeneracy}

In cases of LiDAR degeneracy, we incorporate relative poses from alternative odometry sources, such as visual odometry or learning-based IMU odometry. These relative poses, represented as \([\Delta p_j^i, \Delta q_j^i]\) in local coordinates, are used to constrain the estimated relative position, \(\hat{\alpha}_j^i\), and rotation, \(\hat{\gamma}_j^i\), obtained through joint optimization. To represent the uncertainty of this prior information, we define its covariance matrix as:

\begin{equation}
\mathbf{Cov}_{prior} = \mathbf{I}_{6 \times 6} - \mathbf{Cov},    
\end{equation}

where \(\mathbf{Cov}\) is the \(6 \times 6\) diagonal covariance matrix derived from LiDAR odometry. This matrix provides a normalized measure of observability for each state direction, as introduced in the \textit{Adaptive State Direction} method section.

The relative pose factor, \(\mathbf{e}_{ij}^{prior}\), is formulated as follows:

\begin{equation}
\mathbf{e}_{ij}^{prior} = 
\begin{bmatrix}
\Delta p_j^i \\
\Delta q_j^i
\end{bmatrix}
\ominus
\begin{bmatrix}
\hat{\alpha}_j^i \\
\hat{\gamma}_j^i
\end{bmatrix}.
\end{equation}

Here, the operator \(\ominus\) denotes the difference between the measured relative pose and the estimated relative pose.

This relative pose prior ensures that constraints are evenly distributed across all state directions, thereby mitigating the effects of LiDAR degeneracy. By doing so, it helps prevent numerical instability in the optimization process and improves the robustness of the state estimation framework.

\begin{table*}[!t]
\centering
\caption{\textbf{Learning-based IMU odometry network configuration.}}
\sffamily
\setlength{\tabcolsep}{8pt}
\renewcommand{\arraystretch}{1.15}
\begin{tabular}{|l|l|}
\hline
\multicolumn{2}{|c|}{\textbf{Sampling Specifications}} \\ \hline
Sampling Frequency & 200 Hz \\ 
Window Size & 10 IMU samples \\ 
Input Dimension & \(10 \times 6\) \\ 
Overall Sampling Rate & 20 Hz \\ 
Window Duration & 1 second (200 samples) \\ \hline
\multicolumn{2}{|c|}{\textbf{Data Augmentation}} \\ \hline
Noise Type & Gaussian white noise \\ 
Bias Sampling & Uniform distribution \\ 
Noise Matrices & \(10 \times 200 \times 3\) \\ 
Orientation Augmentation & Random yaw angle rotation \\ \hline
\multicolumn{2}{|c|}{\textbf{Network Setting}} \\ \hline
Framework & PyTorch \\ 
LSTM Time Steps & 10 \\ 
Input Sample Size & \(10 \times 200 \times 6\) \\ 
Supervision Signal & Relative body coordinate velocity \\ 
Optimizer & Adam \\ 
Initial Learning Rate & 0.0001 \\ 
Loss Functions & MSE $\to$ NLL \\ 
Training Epochs & 50 \\ 
Computational Time & 20 hours \\ 
Hardware & NVIDIA Tesla V100 GPU \\ 
Model Selection & Best validation loss \\ \hline
\end{tabular}
\label{tab:network_setting}
\end{table*}

\subsubsection*{\textbf{Implementation Detail on Learning-based IMU Odometry}}
 The network configuration is summarized in \tref{tab:network_setting}.
\paragraph{Sampling Specifications:} 
For the pre-trained IMU model, we employed a sliding window approach, where each window captures 10 seconds of IMU data. Each second consists of 200 frames of IMU samples, resulting in an input dimension of  \(10 \times 200 \times 6\). For IMU data with a frequency lower than 200 Hz, we applied linear interpolation to standardize the frequency at 200 Hz.

\paragraph{Data Augmentation} During the training phase, we introduced random Gaussian white noise and bias into each input sample. Specifically, we sampled a \(1 \times 6\) acceleration and gyroscope bias from a uniform distribution and added two random Gaussian noise matrices with dimensions \(10 \times 200 \times 3\) to each input sample. Additionally, we applied a random yaw angle rotation to enhance the network's learning of yaw angle invariance.

\paragraph{Network Setting} 
The model was implemented in PyTorch and trained using the Adam optimizer with an initial learning rate of 0.0001. The supervision signal was the relative velocity of the sliding window in the body coordinate system. Training spanned 50 epochs and took approximately 20 hours on an NVIDIA Tesla V100 GPU. For further details, please refer to Table S1.

\begin{figure*}[!t]
   \centering
    \includegraphics[width=0.7\textwidth,keepaspectratio]{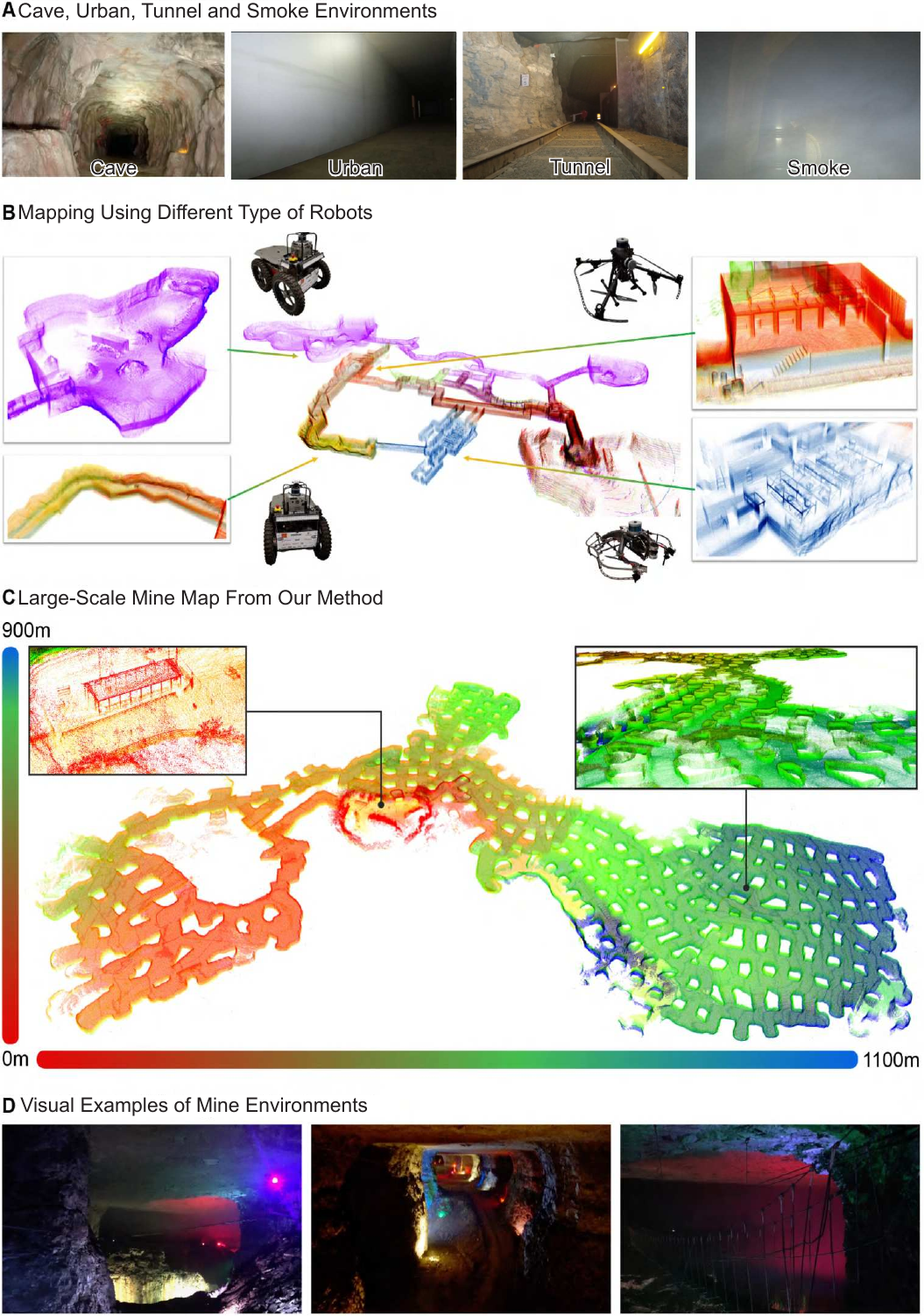}
    \caption{\textbf{Large-scale mapping results from the DARPA subterranean challenge}. \textbf{(A)} The DARPA Final Challenge featured a combination of tunnel, urban, and cave environments \textbf{(B)} Collaborative mapping using both drones and ground vehicles, with different colors representing the maps generated by each vehicle.
    \textbf{(C)} Reconstructed a mine environment during a 1-hour run, covering an area of 990,000 $m^2$ with high precision.
   \textbf{(D)} Visual examples of the mine environments are dark, self-similarity, and lack of distinct features.}
    \label{fig:subt_final}
\end{figure*}

\begin{figure*}[!t]
   \centering
   \includegraphics[width=0.6\textwidth,keepaspectratio]{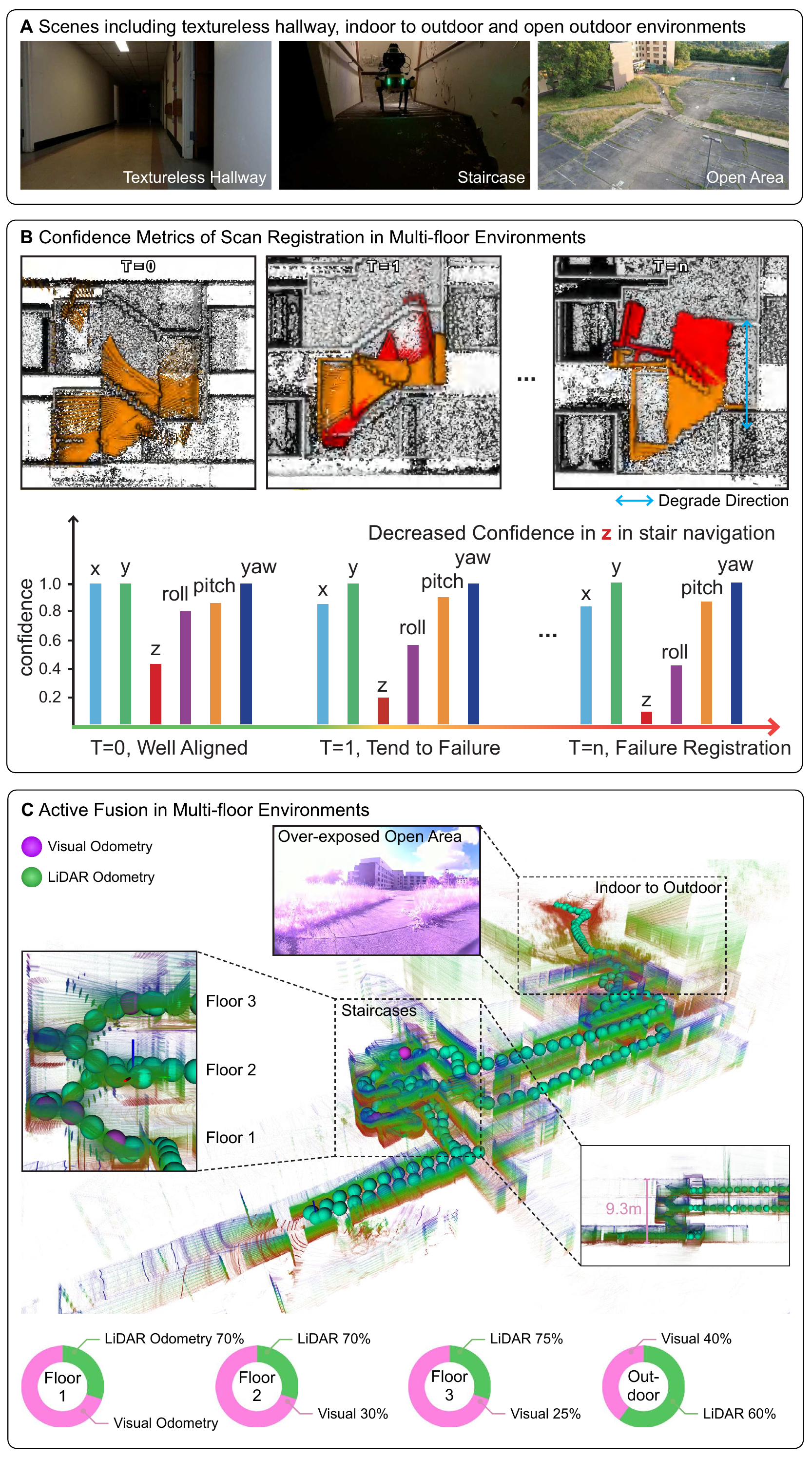}
    \caption{\textbf{Illustration of adaptive state selection and self-weight parameter tuning in a multi-floor environment.} \textbf{(A)} The legged robot navigates featureless corridors, narrow staircases, overexposed areas, and transitions between indoor and outdoor environments. \textbf{(B)}  The system dynamically identifies degraded directions and predicts alignment risks during point cloud registration.
    \textbf{(C)} Our system self-tunes the weight between LiDAR and visual sensors to optimize fusion performance across confined and open spaces. }
    \label{fig:multi_floor}
\end{figure*}

\begin{figure*}[!t]
   \centering
\includegraphics[width=1.0\textwidth,height=1.0\textheight,keepaspectratio]{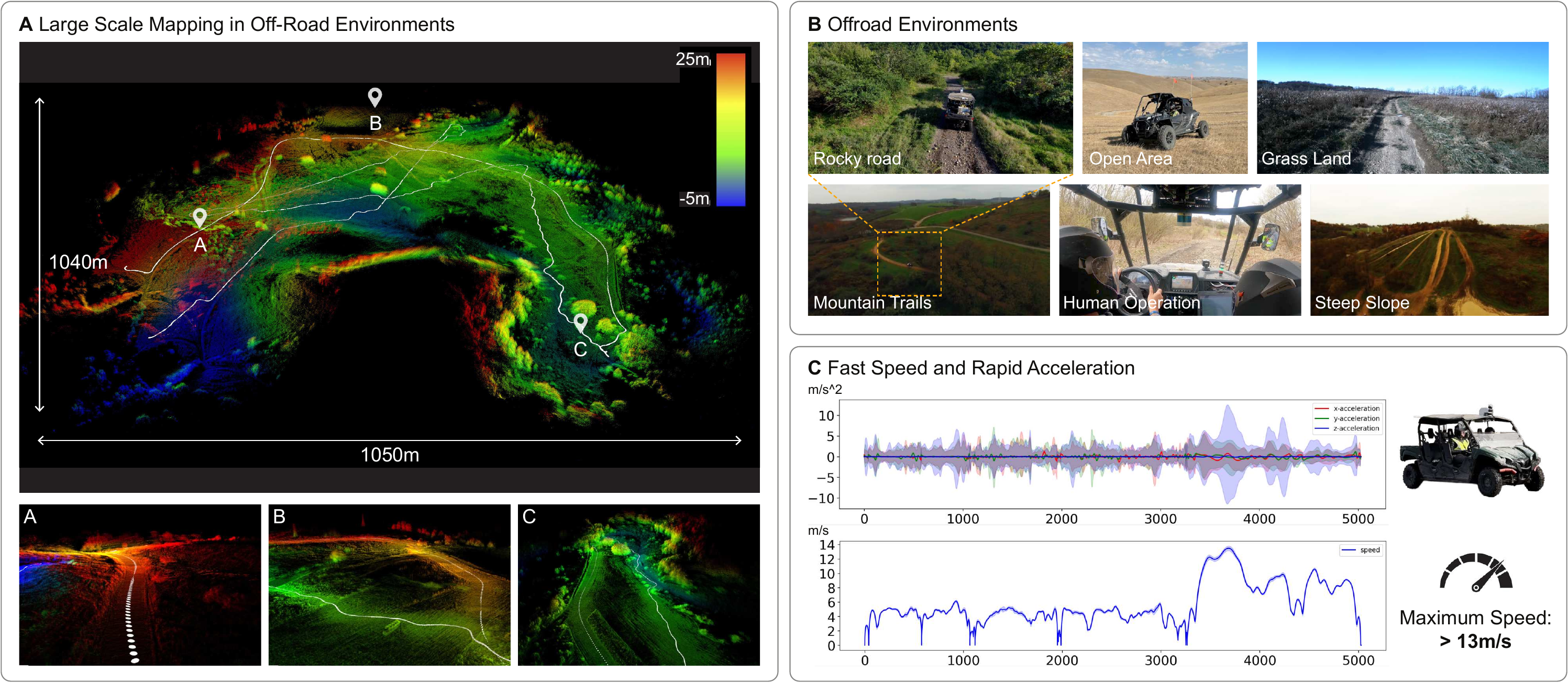}
    \caption{\textbf{Aggressive motion in off-Road environments.} \textbf{(A)} High-fidelity large-scale mapping results. \textbf{(B)} Our ground vehicle traversed uneven off-road surfaces, such as rocks, steep slopes, and unseen ditches, over a 1.38-hour run. \textbf{(C)}  The vehicle reached a maximum speed of \textbf{13 m/s}. Despite these challenges, our method was still able to recover a high-quality map. }
    \label{fig:offroad}
\end{figure*}

\begin{figure*}[!t]
   \centering
    \includegraphics[width=1.0\textwidth,height=\textheight,keepaspectratio]{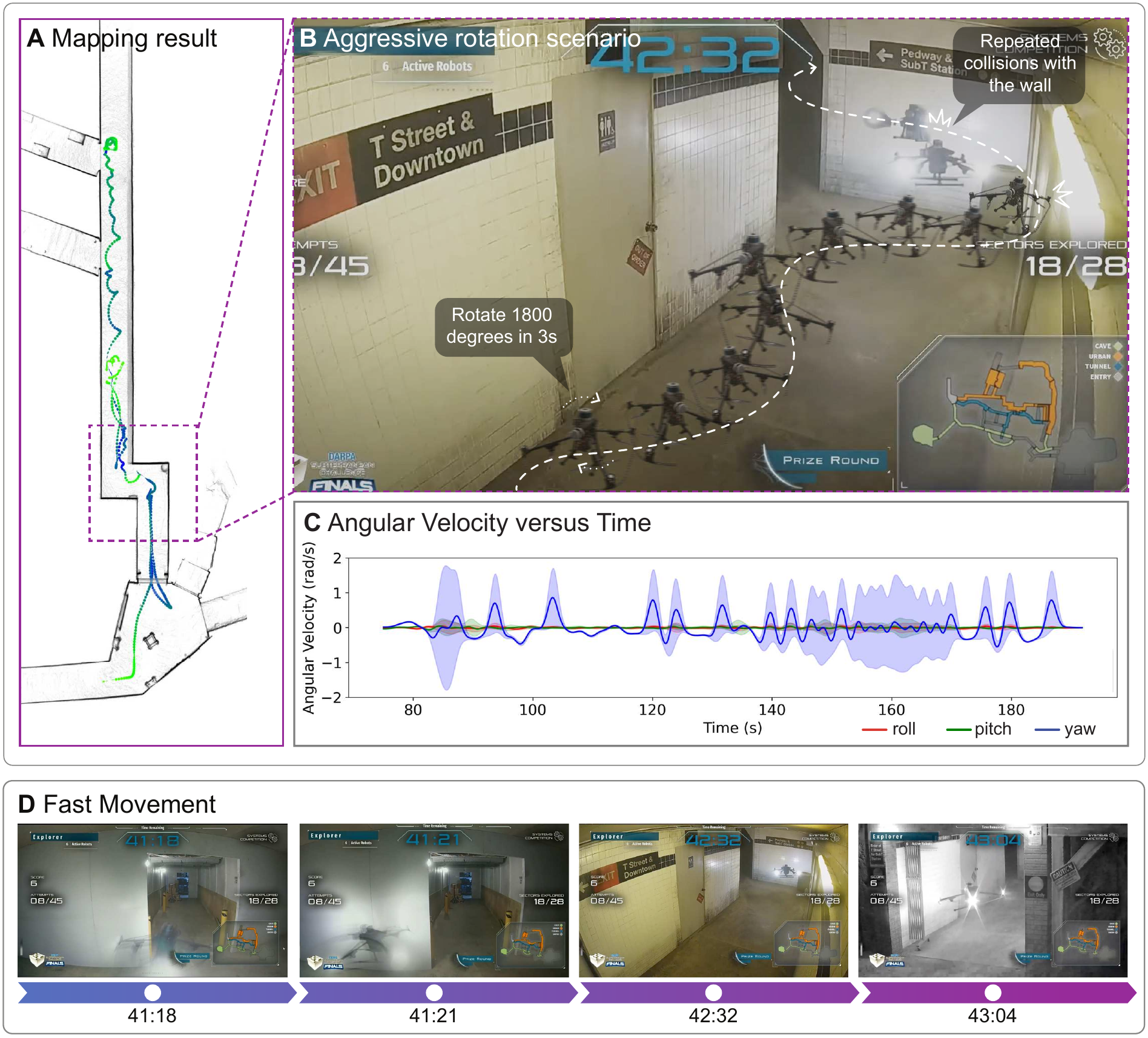}
    \caption{\textbf{Unexpected Collision from a Spinning Drone.} 
    \textbf{(A)} Accurate mapping results despite unexpected collisions. Our odometry solution maintained precise state estimation during a hardware failure, enabling the drone's safe autonomous landing. \textbf{(B)} During the flight, our drone unexpectedly lost control, colliding with walls and spinning erratically for 106 seconds. \textbf{(C)} Angular velocity variations over time. \textbf{(D)} Rapid drone movements in featureless environments.
   }
    \label{fig:subt_drone}
\end{figure*}

\begin{figure*}[!t]
   \centering\includegraphics[width=\textwidth,height=0.8\textheight,keepaspectratio]{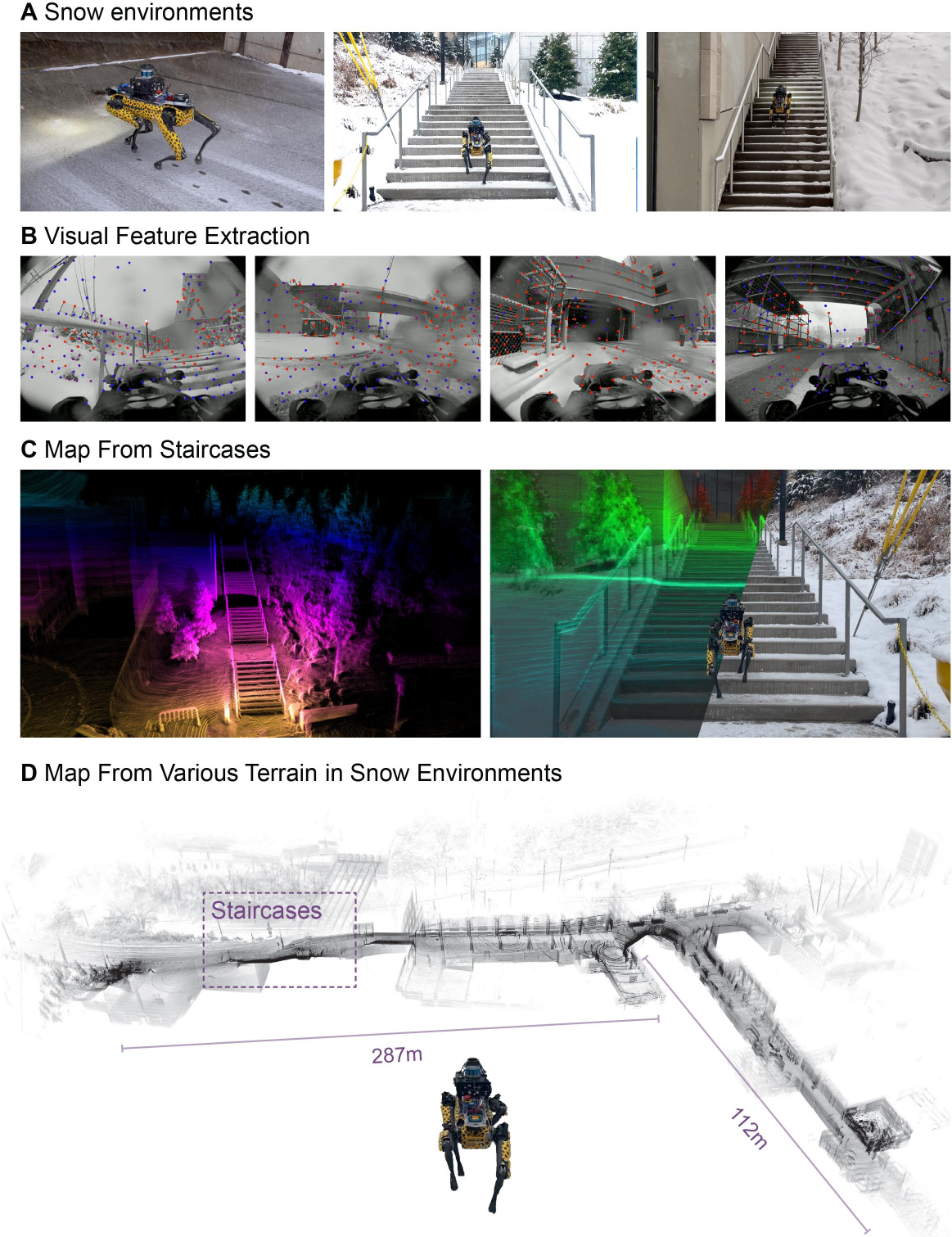}
    \caption{\textbf{Mapping Results in Snow Environments.} \textbf{(A)} Our legged robot navigates steep inclines, slippery surfaces, and staircases, and heavy snow significantly degrades camera systems. \textbf{(B)} Heavy snow leads to degraded image quality for feature detection. \textbf{(C)} The detailed 3D map of staircases. \textbf{(D)} Our method can still recover detailed 3D maps over a distance of 798 meters.}
    \label{fig:snow}
\end{figure*}

\begin{figure*}[!t]
   \centering
    \includegraphics[width=1.0\textwidth,keepaspectratio]{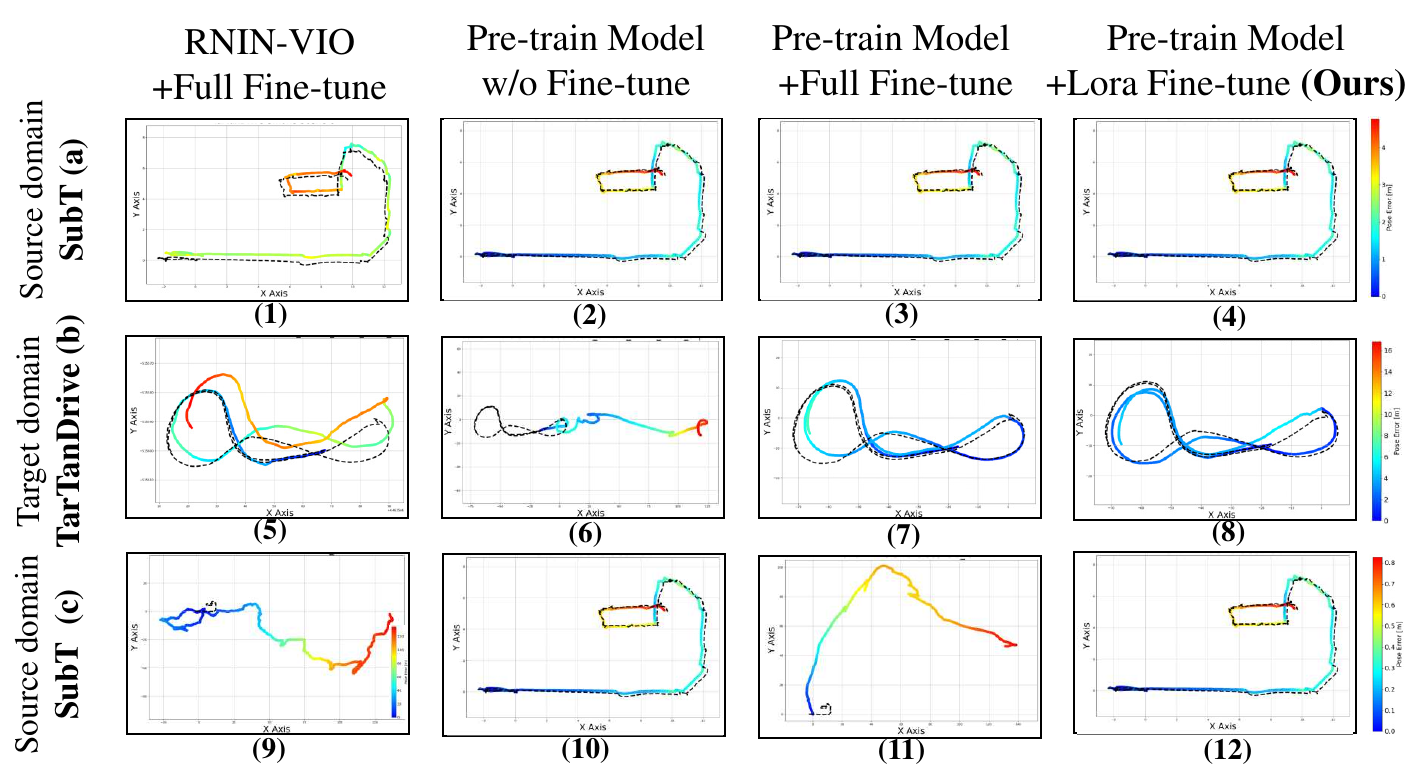}
    \caption{\textbf{Non-forgetting Characteristics of Our Domain Adaption.} We assessed the generalization capabilities of four distinct models across both the same source domain and a target domain. Unlike other models, which exhibit catastrophic forgetting of source domain knowledge following domain adaptation, the LoRA-based approach (last column) effectively mitigates this issue, preserving robust generalization performance. In the visualization, the dashed line represents the ground truth trajectory, while the color-coded lines depict the estimated trajectories.}
    \label{fig:no_forget}
\end{figure*}

\section*{Experiments}
\subsection*{DARPA Subterranean Challenge}
Fig.~\ref{fig:subt_final}A presents results from the DARPA SubT Final Event held in Louisville Mega Cavern, KY. This challenging course featured a combination of tunnel, urban, and cave environments with smoke, dust, low lighting, and featureless surroundings. To improve efficiency, our team deployed a multi-robot system, combining drones and ground robots for collaborative mapping.

Fig.~\ref{fig:subt_final}B highlights the highly accurate state estimation achieved during the SubT Final Challenge. The different colors in the map represent reconstructions from different robots \changed{ (e.g., ground vehicles, aerial drones)}. \changed{These platforms traversed challenging subterranean terrains, covering distances of 596.6 meters, 499.8 meters, 445.2 meters, and 112.0 meters, respectively, over a total runtime of 38 minutes. These large-scale, multi-robot deployments in challenging subterranean environments demonstrate the system's robustness to individual sensor or robot failures.}

Our team's performance was notable across various SubT events. We secured fourth place in the SubT Final Challenge, second place in the Urban Challenge, and first place in the Tunnel Challenge environments. These results emphasize the adaptability and robustness of our odometry system across different environments and robot platforms.

Despite these successes, achieving reliable and efficient performance in large-scale underground environments remains an open challenge, primarily due to the limitations of onboard computation and memory required for autonomous operation. Fig.~\ref{fig:subt_final}C, D shows results from a large mine environment, characterized by darkness and repetitive structures that significantly complicate place recognition using both image and LiDAR data. Despite these challenges, our odometry system accurately reconstructed the environment during a 1-hour run, covering an area of 990,000 $m^2$ with high precision. 

\subsection*{Adaptive State Direction in Multi-floor Environments}  

Multi-floor environments present significant challenges for odometry systems, as most sensors are optimized for horizontal movement and often struggle with vertical transitions. For instance, LiDAR sensors, which are primarily designed for 2D scanning, can encounter difficulties in accurately capturing vertical changes. To address these issues, we implemented our algorithm on a legged robot tasked with navigating through texture-less corridors, dimly lit staircases, and transitions between indoor and outdoor areas, as depicted in Fig.~\ref{fig:multi_floor}A. These challenges are particularly pronounced for both LiDAR-based \cite{shan2020lio, xu2021fast} and visual-based \cite{campos2021orb} solutions.

Despite these obstacles, our sensor fusion system demonstrated exceptional robustness, successfully traversing a total distance of 269.38 meters. This performance underscores its effectiveness in handling complex, real-world scenarios.

\textit{\textbf{Adaptive State Direction:}} Fig.~\ref{fig:multi_floor}B illustrates the confidence metric for scan registration as the legged robot navigates texture-less staircases. The orange scan is the source LiDAR frame, and the red scan is the target frame for alignment. Over time (T=0 to T=n), alignment error increases due to the LiDAR's limited field of view.

The accompanying graph highlights the confidence metric for scan registration. It is apparent that confidence in the Z-direction is substantially lower than in other directions, reflecting insufficient constraints on the robot's pose along the vertical axis. To address this, our system dynamically identifies degradation in confidence and integrates auxiliary sensors as necessary, ensuring accurate and reliable performance under challenging conditions.

\textit{\textbf{Adaptive Weight Parameter Tuning:}} Robust sensor fusion relies on dynamically adjusting the weights between visual and LiDAR sensors to optimize performance. As shown in Fig.~\ref{fig:multi_floor}C, the algorithm identified degraded directions and computed confidence ratios to guide this adjustment. On Floors 1, 2, and 3, the LiDAR confidence values were 30\%, 30\%, and 25\%, respectively, while visual odometry confidence was higher at 70\%, 70\%, and 75\%. However, in outdoor areas, LiDAR confidence in the Z-direction increased to 60\% due to the abundance of geometric features, while visual confidence decreased to 40\%, influenced by overexposed images.

This self-tuning mechanism highlighted the need to adapt sensor contributions to varying environments, ensuring robust performance across diverse scenarios.

\subsection*{Aggressive Motion in Off-Road Environments~}
Aggressive motion poses significant challenges for both visual and LiDAR sensors in odometry systems. Rapid movement can induce motion blur in visual sensors, adversely affecting feature detection and tracking. Similarly, LiDAR sensors are vulnerable to motion distortion due to their sequential sampling, which captures points at different time intervals. When these points are combined into a single frame during fast movement, they create artificial distortions that can compromise the accuracy of localization and mapping.

To evaluate the robustness of our odometry system, we deployed our odometry system on a ground vehicle in off-road environments with aggressive motion. Off-road environments are particularly challenging, often featuring uneven surfaces such as rocks, mud, sand, steep slopes, unseen ditches covered by grass, and open featureless areas as shown in Fig.~\ref{fig:offroad}B.
These terrains can cause falls, drops, collisions,
slippage, high-frequency vibration, and sudden changes in acceleration, making it difficult for odometry systems to maintain accurate localization and mapping, especially when the vehicle is moving at high speeds. 

Fig.~\ref{fig:offroad}C shows the speed and \changed{instantaneous} acceleration of our ground vehicle. The total run lasts 1.38 hours, \changed{and the peak acceleration is around $5\,\mathrm{m/s^2}$}, which is faster than the normal vehicle acceleration ($3m/s^2$). The maximum speed of our vehicle was 13\,m/s. \changed{Despite these challenges}, our odometry system could still reconstruct a high-fidelity large-scale map covering 109200 $m^2$ and recover accurate poses shown in Fig.~\ref{fig:offroad}A.

\subsection*{Unexpected Collision from an Off-nominal Spinning Drone~}

While addressing hardware failures may seem beyond the primary scope of odometry systems, these failures can significantly affect odometry performance. Conversely, odometry systems can play a crucial role in detecting and mitigating sensor and locomotion failures.

During the DARPA SubT Final Challenge, we encountered a real-world example of this interaction. Our drone lost control mid-flight when a propeller blade detached, resulting in erratic behavior. It collided with walls and underwent rapid, uncontrolled rotations, indicating severe locomotion failures (see Fig.~\ref{fig:subt_drone}B). The drone rotated $1800$ degrees in 3s, and this unexpected collision process lasted $106 s$ shown in  Fig.~\ref{fig:subt_drone}D.
Notably, our drone operated in autonomous flight mode, with our odometry output directly feeding into the controller and perception module. If the odometry system cannot withstand a collision for \textbf{$106s$} and provides incorrect poses, which can further compromise the controller's performance. An off-nominal controller can further cause the drone to spin more aggressively, making it increasingly difficult for odometry to recover its pose. This situation can initiate a vicious cycle, potentially leading to a crash.

However, by implementing an adaptive sensor fusion strategy that selectively incorporates features, state direction, and engines, our odometry algorithm successfully overcame this challenge. Consequently, it reconstructed a clear 3D map, as illustrated in Fig.~\ref{fig:subt_drone}A. Our odometry system ultimately secured the drone, allowing it to land safely.

\subsection*{Robustness Experiments in Snow Environments} To evaluate the robustness of our state estimation algorithm, we conducted tests with legged robots in challenging snowy environments. The robots navigated steep inclines, slippery surfaces, and staircases (Fig.~\ref{fig:snow}A). These conditions are particularly demanding, as slipping and compressible terrain can disrupt locomotion patterns, negatively affecting state estimation. Additionally, heavy snow can degrade camera systems, leading to reduced image quality and potentially introducing rain streaks and ice into images (Fig.~\ref{fig:snow}B). Snow also impacts the quality of LiDAR scans \cite{dreissig2304survey}.

Despite these obstacles, our method successfully generated a high-fidelity 3D map under heavy snow conditions. This included mapping steep slopes, staircases, and transitions from indoor to outdoor environments over a distance of 798 meters (Fig.~\ref{fig:snow}D). The detailed 3D map of staircases (Fig.~\ref{fig:snow}C) demonstrates our system's precision and robustness.

\subsection*{Non-forgetting Characteristics of Our Domain Adaptation}
We evaluated the performance of four models in both the same source domain and a target domain. These models included the SOTA expert model (RNIN-VIO)\cite{chen2021rnin}, the pre-trained IMU model without fine-tuning, the fully fine-tuned IMU model, and IMU model fine-tuned with parameter-efficient low-rank adapter (LoRA) \cite{hu2022lora} shown in Fig.~\ref{fig:no_forget}.

In the first row \fref{fig:no_forget}(a), we presented the results from different models tested on the same source domain (i.e., SubT-Dataset) shown in \fref{fig:no_forget}(1)-(4). Our findings reveal that the pre-trained IMU model achieves more accurate trajectory estimations than the expert model in the source domain, highlighting its strong generalization capabilities.

The second row \fref{fig:no_forget}(b) illustrates the performance of these models in the TartanDrive environment, an entirely unseen target domain representing a distribution outside the pre-training dataset. As expected, the pre-trained model without fine-tuning produces the poorest results. While the RNIN-VIO model with full fine-tuning achieves reasonable trajectory estimations, it suffers from relatively large pose errors. In contrast, integrating a LoRA module into the pre-trained IMU-based model significantly enhances its performance, achieving similar accuracy with the fully fine-tuned IMU-based model, as shown in \fref{fig:no_forget}(7)-(8).

In the third row \fref{fig:no_forget}(c), we assessed the performance of the aforementioned models on the source domain again after target domain adaptation. Both the RNIN-vio model with fully fine-tuned  \fref{fig:no_forget}(9) and the pre-trained model with fully fine-tuned \fref{fig:no_forget}(11) perform poorly, illustrating the phenomenon of catastrophic forgetting. In contrast, our LoRA-based fine-tuning design maintains accurate estimation results in the source domain. This demonstrates that the LoRA-based method effectively mitigates forgetting, enabling robust application across diverse scenarios while maintaining strong generalization performance.

% If your supplement is very short you might need to uncomment the following line to avoid
% layout problems with the figures and tables.
%\newpage

%%%%%%%%%%%%%%%% SUPPLEMENTARY FIGURES %%%%%%%%%%%%%%%

% \begin{figure} % Do not use \begin{figure*}
% 	\centering
% 	\includegraphics[width=0.6\textwidth]{example_figure} % for an image file named example_figure.*
% 	% Pick an appriopriate width for the size of the image

% 	% Captions go below figures
% 	\caption{\textbf{All captions must start with a short bold sentence, acting as a title.}
% 		Follow the same style as main text figures.
% 		If the design is substantially the same as another figure, avoid repeating information
% 		e.g. say `Same as Figure~\ref{fig:example}, but for the control sample.'}
% 	\label{fig:sup_example} % give each figure a logical label name
% \end{figure}

%%%%%%%%%%%%%%%% SUPPLEMENTARY TABLES %%%%%%%%%%%%%%%

% \begin{table} % Do not use \begin{table*}
% 	\centering
% 	% Captions go above tables
% 	\caption{\textbf{All captions must start with a short bold sentence, acting as a title.}
% 		Follow the same style as main text tables.
% 		If the design is similar to previous tables, avoid repetition by refering back to them.}
% 	\label{tab:sup_example} % give each table a logical label name

% 	\begin{tabular}{lccr} % four columns, alignment for each
% 		\\
% 		\hline
% 		A & B & C & D\\
% 		\hline
% 		1 & 2 & 3 & 4\\
% 		2 & 4 & 6 & 8\\
% 		3 & 5 & 7 & 9\\
% 		\hline
% 	\end{tabular}
% \end{table}

%%%%%%%%%%% CAPTIONS FOR OTHER SUPPLEMENTARY FILES %%%%%%%%%%

\clearpage % Clear all remaining figures and tables then start a new page

% \paragraph{Caption for Movie S1.}
% \textbf{All captions must start with a short bold sentence, acting as a title.}
% Then explain what is shown in the supplementary video file.
% Give as much detail as you would for a figure e.g. explain axes, color maps etc.
% If the video is an animated equivalent of one of the static figures, state e.g.
% `Animated version of Figure~\ref{fig:example}.'

% \paragraph{Caption for Data S1.}
% \textbf{All captions must start with a short bold sentence, acting as a title.}
% Then explain what is included in the supplementary data file.
% Give as much detail as you would for a table e.g. explain the meaning of every column,
% units used, any special notation etc.

%%%%%%%%%%%%%%%% SUPPLEMENTARY REFERENCES %%%%%%%%%%%%%%%

% Do NOT include a reference list in the supplement.
% All references must be in a single list at the end of the main text.
% The copyeditors will ensure that the correct reference list appears with each version of the paper
% (print, HTML, PDF, mobile app, metadata for bibliographic databases etc.)

\end{document}